\documentclass[11pt]{article}
\usepackage[margin=0.8in]{geometry}

\usepackage{amsmath}
\usepackage{amssymb}
\usepackage{booktabs}
\usepackage{graphicx}
\usepackage{xcolor} 
\usepackage{caption}
\usepackage{microtype}
\usepackage{tabularx}
\usepackage{array}
\usepackage{makecell}
\usepackage[numbers,sort&compress]{natbib}
\usepackage{ragged2e}
\usepackage[hidelinks]{hyperref}

\graphicspath{{Figure/}}
\newcolumntype{L}[1]{>{\RaggedRight\arraybackslash}p{#1}}
\newcolumntype{M}[1]{>{\centering\arraybackslash}m{#1}}
\newcolumntype{Y}{>{\RaggedRight\arraybackslash}X}
\renewcommand{\arraystretch}{1.32}
\newcommand{\affil}[1]{\begin{center}\small #1\end{center}}
\newcommand{\keywords}[1]{\medskip\noindent\small\textbf{Keywords:} #1\par\medskip}

\makeatletter
\newcommand{\corrfootnote}[1]{%
  \begingroup
    \renewcommand{\thefootnote}{*}%
    \def\Hy@footnote@currentHref{corrauthors}
    \footnotetext{#1}%
  \endgroup
}
\makeatother

\hypersetup{
  pdftitle={Equation Recast for Canonical Operator Learning Across Parametric PDEs},
  pdfauthor={Qiyun Cheng, Valentin Duruisseaux, Cesar F. Clauser, Md Hossain Sahadath,
             Huihua Yang, Shaowu Pan, Nathaniel Ferraro, Anima Anandkumar,
             Wei Ji, Cristina Rea}
}

\begin{document}
\justifying

\title{\bfseries Equation Recast for Canonical Operator Learning Across Parametric PDEs}

\author{Qiyun Cheng$^{1,*}$, Valentin Duruisseaux$^{2}$, Cesar F. Clauser$^{1}$,
Md Hossain Sahadath$^{3}$, Huihua Yang$^{1}$,\\
Shaowu Pan$^{3}$, Nathaniel Ferraro$^{4}$, Anima Anandkumar$^{2}$,
Wei Ji$^{3,*}$, and Cristina Rea$^{1,*}$}
\date{}
\maketitle

\affil{$^1$Plasma Science and Fusion Center, Massachusetts Institute of Technology, Cambridge, MA, USA}
\affil{$^2$Computing and Mathematical Sciences, California Institute of Technology, Pasadena, CA, USA}
\affil{$^3$Department of Mechanical, Aerospace, and Nuclear Engineering, Rensselaer Polytechnic Institute, Troy, NY, USA}
\affil{$^4$Princeton Plasma Physics Laboratory, Princeton, NJ, USA}

\corrfootnote{Corresponding authors:
  \texttt{chengq@psfc.mit.edu}, \texttt{jiw2@rpi.edu}, and \texttt{crea@psfc.mit.edu}}

\begin{abstract}
\noindent
Learning solution operators across broad parameter ranges can require substantial coverage of both input functions and physical parameters, particularly for purely data-driven parametric models. In addition, the resulting models may fail silently outside the training distribution.
We introduce equation recast, which reformulates parametric operator learning as the learning of a single canonical operator.
Parameter-induced operator variations are derived analytically from the governing equation and absorbed into effective sources, enabling zero-shot prediction across new parameter regimes.
Across multi-parameter, nonlinear, and singular PDE settings, equation recast supports extrapolation, integrates sparse heterogeneous datasets in a shared canonical representation, and uses loss of convergence as an internal warning signal for failure of the recast iteration.
In high-fidelity tokamak simulations for nuclear fusion, the framework unifies electron-temperature data across four device geometries through canonical-domain mapping within one jointly trained operator.
Equation recast provides a route toward reusable neural PDE solvers combining equation-guided transfer, data efficiency, and monitorable inference.
\end{abstract}


\section{Introduction}
\label{sec:intro}

Some of the most consequential applications of scientific machine learning, from fusion-plasma control to real-time digital twins, depend on surrogates for partial differential equations (PDEs) whose governing operators vary with physical parameters~\cite{karniadakis2021piml, azizzadenesheli2024neuraloperators, niederer2021digitaltwin}. 
A useful surrogate must therefore generalize not only across solutions of a fixed PDE, but also across a family of related PDEs that share a governing form. 
This remains difficult because high-fidelity simulations are expensive, parameter configurations are often sampled sparsely and unevenly, and neural networks may produce apparently plausible predictions in untested regimes. 
Reliable parametric PDE learning consequently depends on how parameter variation is represented, how effectively limited data cover the relevant regimes, and whether unreliable predictions can be identified before they are used.

Existing approaches address parts of this problem, but leave important gaps. 
The first category is direct parametric learning, in which a model maps inputs, including sources, parameters and sometimes geometry, directly to solutions. 
Physics-informed neural networks (PINNs) embed the governing equations into the training objective with physical parameters as additional inputs~\cite{raissi2019pinn,karniadakis2021piml}, and have shown strong performance with limited or no data~\cite{raissi2020hiddenfluid, wang2022neutronpinn, xiang2025pinnimplicitflow, wu2025fuzzypinn, grossmann2024pinnvsfem}. 
Operator learning constructs mappings between function spaces and have demonstrated strong generalization across many parametric PDE settings, although performance can depend on the coverage and representation of the training distribution~\cite{azizzadenesheli2024neuraloperators, kovachki2023neuraloperator, li2020gkn, li2021fno, lu2021deeponet, cheng2025fourierdeeponet, lu2022comparison, venturi2023svddeeponet,cheng2026jcp}. 
Physics-informed operator learning combines PDE-residual losses with this function-space generality~\cite{goswami2022pivdeeponet, wang2021pideeponet, li2024pino, ganeshram2025fcpinohighprecisionphysicsinformed, lin2025mGNO}.
Related approaches also use learned operators as iterative or preconditioning components inside classical solvers~\cite{chen2025gnnpreconditioner, kopanicakova2025deeponetprecond, li2025neuralprecondop}. 
Across these formulations, parameter variation is typically carried by input channels, conditioning variables, geometry encodings, residual constraints or the sampled training distribution. 
Reliable performance can therefore remain sensitive to which parameter configurations are represented during training or physics-informed optimization, which is limiting when high-fidelity data are sparse~\cite{zhu2023reliableextrapolation, subramanian2023foundation}. 
The second category is transfer learning and adaptation, including fine-tuning and meta-learning, which reuse models across configurations~\cite{goswami2020transferpinn, wen2022ufno, psaros2022metapinnloss, penwarden2023metalearning, qin2022metapde, zhang2023metano}. 
These methods can transfer models between discrete regimes, but they still treat generalization primarily as a statistical adaptation problem and do not directly address continuous parameter variation under limited data coverage. 
The third category comprises classical perturbation, continuation and reduced-basis methods, which propagate solutions across parameter variation using analytical structure such as Taylor expansion, branch following or affine parametric decomposition~\cite{keller1987numerical, allgower2003continuation, quarteroni2016reducedbasis}. 
Closely related fixed-point and defect-correction schemes repeatedly apply an approximate or frozen operator inverse while feeding back the residual against the target operator~\cite{stetter1978defect, keller1987numerical, allgower2003continuation}. These methods provide the numerical precedent for iterative correction and are typically formulated as problem-specific numerical procedures rather than reusable learned surrogates.

\begin{figure}[htbp]
    \centering
    \includegraphics[width=1.0\textwidth]{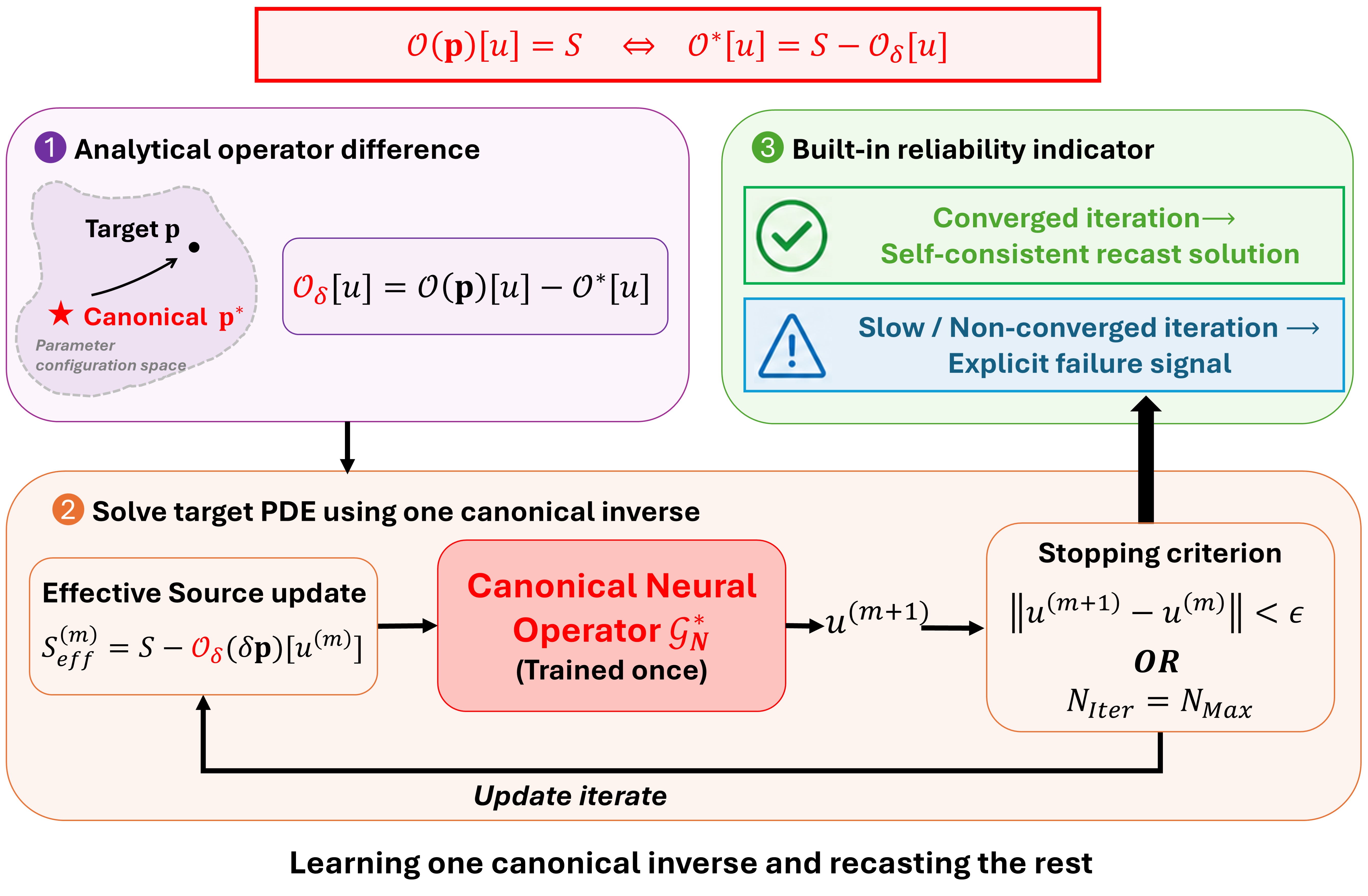}
    \caption{\textbf{Equation recast reduces parametric operator learning to a single canonical inverse.}
    For a target PDE \(\mathcal{O}(\mathbf{p})[u]=S\), a reference configuration \(\mathbf{p}^*\) defines the canonical operator \(\mathcal{O}^*=\mathcal{O}(\mathbf{p}^*)\).
    \textbf{(1)} For a target parameter configuration \(\mathbf{p}\), the operator difference \(\mathcal{O}_{\delta}[u]=\mathcal{O}(\mathbf{p})[u]-\mathcal{O}^*[u]\) is derived analytically from the governing equation.
    \textbf{(2)} A canonical neural operator \(\mathcal{G}_N^*\approx(\mathcal{O}^*)^{-1}\) is trained once at the reference configuration. 
    During inference, the effective source \(S_{\mathrm{eff}}^{(m)}=S-\mathcal{O}_{\delta}[u^{(m)}]\) is updated iteratively and repeatedly passed through the fixed canonical operator until convergence.
    \textbf{(3)} Convergence yields a self-consistent recast solution, whereas slow or non-convergent iteration signals loss of stable continuation from the chosen canonical operator.
    Parameter dependence therefore enters inference analytically through the governing equation while the learned operator remains fixed.}
    \label{fig:Fig1}
\end{figure}

In this work, we introduce a new operator-learning framework, termed equation recast, that addresses these gaps by changing the object to be learned.
As illustrated in Figure~\ref{fig:Fig1}, rather than approximating an entire parameter-indexed family of inverse operators, it learns a single canonical inverse operator defined at a fixed reference parameter configuration.
For any target configuration, the parameter-induced operator difference between the target governing operator and the canonical operator is derived analytically from the known equation and absorbed into an effective source.
This reformulation preserves the original governing equation while separating analytically known parameter variation from the learned canonical inverse.
During inference, the effective source is updated using the current solution estimate and repeatedly passed through the same canonical neural operator, forming a fixed-point iteration to predict the solution of the target PDE.
Equation recast therefore enables parameter generalization through analytical reformulation rather than requiring the learned model to infer parameter dependence entirely from sampled configurations.
Sparse and heterogeneous data generated at scattered parameter configurations can be recast into the same canonical representation, enriching the effective-source space without dense sampling of the parameter space.
The iterative inference procedure provides an internal convergence diagnostic.
When the iteration converges, it yields a self-consistent recast prediction whose accuracy also depends on the learned canonical inverse.
Near singularities, resonances, or other regimes where the recast iteration becomes unstable, slow convergence or divergence can signal that the chosen canonical operator no longer provides a reliable iterative solution pathway.
This makes certain failures observable through the inference process rather than leaving them hidden in a direct surrogate prediction.
Because the recast is formulated at the equation level, it remains independent of the particular neural-operator architecture and can be combined with different operator-learning or physics-informed approaches.

Equation recast is evaluated across a sequence of settings designed to isolate its main properties, as summarized in Table~\ref{tab:equation_recast_cases}, including multi-parameter extrapolation, sparse-data efficiency, behavior near singular regimes, and data-efficient extrapolation in nonlinear multiscale dynamics.
We further evaluate the framework in high-fidelity tokamak energy-equation simulations.
These simulations involve complex, time-dependent coefficient and source fields, with data drawn from four tokamak device geometries.
To combine these datasets, each physical domain is mapped to a shared canonical domain, where geometry enters the transformed PDE through known Jacobian-dependent coefficients \cite{hughes2000fem}.
These terms are incorporated into the effective source together with the plasma-coefficient corrections, allowing data from all four device geometries to be used for joint training.
The results demonstrate that equation recast can be combined with domain canonicalization to unify heterogeneous multi-device data in a realistic fusion application.

This work makes three principal contributions.
First, equation recast separates analytically known operator variation from the learned inverse, allowing one canonical operator to be reused across parameter configurations.
Second, it provides a common representation for sparse and heterogeneous datasets generated at different parameter values, improving source-space coverage without requiring dense parameter sampling.
Third, it converts inference into a monitorable iterative procedure whose convergence behavior exposes important limits of extrapolation.
These contributions position equation recast as a framework for reusable neural PDE solvers in which parameter generalization is supported by canonical operator learning and analytical equation structure, rather than by broad empirical coverage alone.

The remainder of this paper is organized as follows.
Section~2 presents the equation recast formulation, including its fixed-point consistency, extension to time-dependent problems, and geometry canonicalization.
Section~3 evaluates the framework across representative PDE benchmarks and high-fidelity multi-device tokamak simulations.
Section~4 discusses the main findings, limitations, and broader implications of the proposed framework.
Additional methodological details and numerical results are provided in the Appendices.

\begin{table*}[t]
\centering
\caption{\textbf{Benchmark PDE settings for evaluating equation recast.}
Each problem is formulated around a reference configuration defining a canonical operator.
For the steady problems, parameter variation enters through an analytically known operator difference
\(\mathcal{O}_{\delta}(\delta\mathbf{p})[u]\), which is absorbed into the effective source while the canonical inverse remains fixed.
In the Tokamak application, geometry is first mapped to a canonical domain and enters the transformed equation through Jacobian-dependent metric terms \(\mathrm{K}\). The time-dependent equation is then discretized and recast for the temperature increment about a fixed reference state (coefficients \(\tilde{\mathbf{p}}^*\), geometry \(\mathrm{K}^*\), temperature field \(\widetilde{T}_e^{*}\), and time step \(\Delta t^*\)), so that the canonical object is again the inverse of a fixed operator, the one-step increment operator \(\widetilde{\mathcal{O}}^{*}\) of the reference state, and parameter-, geometry-, and state-induced differences are incorporated into its effective source.}
\label{tab:equation_recast_cases}

\vspace{4pt}
\footnotesize

\begin{tabularx}{\textwidth}{@{}
L{0.15\textwidth}
L{0.28\textwidth}
M{0.12\textwidth}
Y
@{}}
\toprule
\makecell[l]{\textbf{Objective}} &
\makecell[l]{\textbf{PDE}} &
\makecell[c]{\textbf{Canonical}\\[-1pt]\textbf{config.}} &
\makecell[l]{\textbf{Recast Operator Form}} \\
\midrule

Multi-parameter extrapolation
&
\makecell[tl]{Advection--diffusion--reaction (1D)\\[4pt]
$-u''+\mathrm{Pe}\,u'+\mathrm{Da}\,u=S$}
&
\makecell[c]{$\mathrm{Pe}^*=4$\\[6pt]
$\mathrm{Da}^*=2$}
&
\makecell[tl]{$\mathcal{O}^*[u]
= S-(\mathrm{Pe}-\mathrm{Pe}^*)u'-(\mathrm{Da}-\mathrm{Da}^*)u$\\[4pt]
$\mathrm{Pe}\in[1,20],\qquad \mathrm{Da}\in[1,20]$}
\\[4pt]

\midrule

Sparse heterogeneous data utilization
&
\makecell[tl]{Reaction-Diffusion (1D)\\[4pt]
$-u''+ku=S$}
&
\makecell[c]{$k^*=0.785$\\[6pt]
$k_{\mathrm{new}}=0.05$}
&
\makecell[tl]{$\mathcal{O}^*[u]
= S-\bigl(k-k^*\bigr)u$\\[4pt]
$k\in[0.01,10]$}
\\[4pt]

\midrule

Identification of failures
&
\makecell[tl]{Helmholtz equation (1D)\\[4pt]
$u''+k^2u=-S$}
&
\makecell[c]{$k^*=0.785$}
&
\makecell[tl]{$\mathcal{O}^*[u]
= -S-\bigl(k^2-(k^*)^2\bigr)u$\\[4pt]
$k\in(0.628,0.942)\cup(0.942,1.257)$ \\ [4pt] 
near resonance regimes}
\\[4pt]

\midrule

Nonlinear multiscale operator
&
\makecell[tl]{Navier-Stokes, vorticity form (2D)\\[4pt]
$\mathbf{u}\cdot\nabla\omega-\dfrac{1}{\mathrm{Re}}\Delta\omega=S$}
&
\makecell[c]{$\mathrm{Re}^*=250$}
&
\makecell[tl]{$\mathcal{O}^*[\omega]
= S+\left(\dfrac{1}{\mathrm{Re}}-\dfrac{1}{\mathrm{Re}^*}\right)\Delta\omega$\\[12pt]
$\mathrm{Re}\in[50,400]$}
\\[4pt]

\midrule

Parameter and geometry variation unification
&
\makecell[tl]{Electron temperature (energy)\\ equation of the MHD system (2D)}
&
\makecell[c]{$\tilde{\mathbf{p}}^*,\,\mathrm{K}^*,$\\[2pt]$\widetilde{T}_e^{*},\,\Delta t^*$}
&
\makecell[tl]{$\widetilde{\mathcal{O}}^{*}[\Delta\widetilde{T}_e^{\,n+1}]
= \widetilde{Q}_{\mathrm{tot}}^{\,n+1} - \widetilde{\mathcal{O}}_{\delta}[\Delta\widetilde{T}_e^{\,n+1}]$\\[4pt]
$(\widetilde{T}_e^{\,n}, \widetilde{S}_{\mathrm{eff}}^{\,n+1}) \longmapsto \Delta\widetilde{T}_e^{\,n+1}$\\[4pt]
$\tilde{\mathbf{p}},\,\mathrm{K}$ across multiple devices}
\\[4pt]
\bottomrule
\end{tabularx}
\end{table*}

\clearpage
\section{Method}
\label{sec:method}

\subsection{Equation recast formulation}
\label{sec:method:recast}

Consider a family of partial differential equations sharing the same governing form but defined under different parameter configurations, written abstractly as
\begin{equation}
\mathcal{O}(\mathbf{p})[u] = S,
\label{eq:parametric_pde}
\end{equation}
where $u$ denotes the solution field, $S$ denotes the source or forcing term, and $\mathbf{p}$ is a possibly high-dimensional parameter vector specifying the PDE instance, for example, through material properties, transport coefficients, constitutive parameters, or dimensionless numbers.
Variation in $\mathbf{p}$ induces variation in the governing operator, so that the learning problem involves a family of related operators rather than a single PDE instance.
A common formulation of parametric PDE learning is to directly approximate the corresponding solution operator as a parameter-dependent mapping:
\begin{equation}
\mathcal{G}_N(\mathbf{p}) \approx \mathcal{O}(\mathbf{p})^{-1},
\end{equation}
so that the solution at a target parameter configuration is obtained as
\begin{equation}
u = \mathcal{G}_N(\mathbf{p})[S].
\end{equation}
The learned model must therefore approximate an entire family of inverse operators across the coupled space of source functions and parameter configurations.

Equation recast adopts a different formulation, in which only a canonical inverse operator at a fixed reference configuration is learned, while parameter variation is handled analytically through the governing equation.
Let $\mathbf{p}^*$ be a reference parameter and define the canonical operator
\begin{equation}
\mathcal{O}^* \equiv \mathcal{O}(\mathbf{p}^*).
\label{eq:canonical_operator}
\end{equation}
For a target parameter $\mathbf{p}$, define the operator difference
\begin{equation}
\mathcal{O}_\delta(\delta \mathbf{p})[u]
= \mathcal{O}(\mathbf{p})[u] - \mathcal{O}^*[u],
\qquad
\delta \mathbf{p} = \mathbf{p} - \mathbf{p}^*.
\label{eq:operator_difference}
\end{equation}
Substituting Eq.~\eqref{eq:operator_difference} into Eq.~\eqref{eq:parametric_pde} yields the recast form
\begin{equation}
\mathcal{O}^*[u] = S - \mathcal{O}_\delta(\delta \mathbf{p})[u].
\label{eq:recast_equation}
\end{equation}
This naturally defines an effective source
\begin{equation}
S_{\mathrm{eff}} = S - \mathcal{O}_{\delta}(\delta \mathbf{p})[u],
\end{equation}
in which parameter-dependent variation is absorbed as a structured additional source term.
Let $\mathcal{G}^*_N \approx (\mathcal{O}^*)^{-1}$ denote the learned inverse of the canonical operator.
Starting from an initial estimate $u^{(0)}$, the solution of the target PDE is predicted by repeatedly applying the learned canonical operator to the updated effective source:
\begin{equation}
u^{(m+1)} = \mathcal{G}^*_N\!\left[S_{\mathrm{eff}}^{(m)}\right],
\qquad
S_{\mathrm{eff}}^{(m)}
= S - \mathcal{O}_{\delta}(\delta\mathbf{p})[u^{(m)}],
\qquad m = 0,1,2,\dots.
\label{eq:recast_iteration}
\end{equation}

Equation recast separates the learning of the canonical inverse from analytically known parameter variation.
This reduces the parameterization burden placed on the learned model, allows prior PDE structure to be incorporated directly, and enables extrapolation through an analytically structured recast procedure rather than requiring the learned model to represent the full parameter dependence directly. 
The iterative procedure is closely related to classical defect-correction and refinement methods~\cite{stetter1978defect}, but employs a learned canonical inverse that can be reused across parameter and geometry configurations without retraining for each problem instance.

The formulation of equation recast implicitly assumes that, within the parameter regime of interest for a given PDE, for a given source $S$, finite variations in the parameter $\mathbf{p}$ lead to correspondingly bounded variations in the solution $u$. 
Thus, equation recast does not guarantee global convergence across parameter space.
The Helmholtz example illustrates how an operator singularity can terminate stable continuation from a chosen canonical configuration.
The useful regime is further determined by the regularity of the target operator, the approximation quality of the learned canonical inverse, and the convergence of the resulting recast iteration.

Equation recast is particularly relevant to applications requiring rapid and repeated PDE evaluations, such as optimization, design exploration, and real-time control.
These applications often explore variations around a baseline design or nominal operating condition, providing a natural guideline for selecting the canonical parameter $\mathbf{p}^*$.
The reference configuration can therefore be chosen as the baseline, nominal, or most frequently encountered condition of interest.
From this reference point, equation recast enables a single trained canonical operator to be reused across a range of related conditions, potentially reducing the need for large-scale data generation through exhaustive parameter scanning while explicitly incorporating the governing physics during extrapolative prediction.

The recast identity is formulated at the equation level before discretization and can therefore be paired with different numerical discretizations and neural-operator architectures.
In this study, we implement the canonical inverses using models from the \textit{NeuralOperator} library~\cite{kossaifi2025librarylearningneuraloperators,duruisseaux2025guide}, with the Fourier Neural Operator~\cite{li2021fno,duruisseaux2025guide} as the primary learner.

\subsection{Time-dependent formulation}
\label{sec:method:timedep}

For time-dependent PDEs, equation recast is applied to a one-step surrogate operator that approximates the solution evolution over a single time step.
Consider
\begin{equation}
\frac{\partial u}{\partial t} + \mathcal{N}(\mathbf{p})[u] = S,
\label{eq:timedep_pde}
\end{equation}
where $\mathcal{N}(\mathbf{p})$ is the spatial operator, which may act nonlinearly on $u$, and $S$ is a source without explicit dependence on $u$.
Let $u^n$ denote the field at time $t^n$, $\Delta t^{n+1}=t^{n+1}-t^n$ be the corresponding time step, and define the solution increment
\begin{equation}
\Delta u^{n+1} \equiv u^{n+1}-u^n.
\label{eq:increment_def}
\end{equation}
Backward-Euler discretization gives $\Delta u^{n+1}/\Delta t^{n+1}+\mathcal{N}(\mathbf{p})[u^n+\Delta u^{n+1}]=S^{n+1}$, where $S^{n+1}$ and $\mathbf{p}$ are evaluated at $t^{n+1}$.
In practice, learning the increment rather than the full field is usually advantageous when the change over one step is small compared with the background field.
Taking the increment as the unknown, the discretized equation becomes
\begin{equation}
\mathcal{O}_{\mathrm{inc}}(\mathbf{p},\Delta t^{n+1};u^n)
\!\left[\Delta u^{n+1}\right]
\equiv
\frac{\Delta u^{n+1}}{\Delta t^{n+1}}
+\mathcal{N}(\mathbf{p})
\!\left[u^n+\Delta u^{n+1}\right]
=S^{n+1}.
\label{eq:increment_operator}
\end{equation}

This has the same operator form as Eq.~\eqref{eq:parametric_pde}, allowing equation recast to be applied directly.
Here, the time step $\Delta t^{n+1}$ and the current field $u^n$ act as additional parameters conditioning the operator.
Choosing fixed reference values $\mathbf{p}^*$, $\Delta t^*$, and $u^*$ therefore defines a single canonical increment operator
\begin{equation}
\mathcal{O}^*\!\left[\Delta u\right]
\equiv
\mathcal{O}_{\mathrm{inc}}(\mathbf{p}^*,\Delta t^*;u^*)
\!\left[\Delta u\right]
=
\frac{\Delta u}{\Delta t^*}
+\mathcal{N}(\mathbf{p}^*)
\!\left[u^*+\Delta u\right].
\label{eq:canonical_increment_operator}
\end{equation}
For each target time step, the operator difference is
\begin{equation}
\mathcal{O}_\delta\!\left[\Delta u\right]
=
\left(\frac{1}{\Delta t^{n+1}}-\frac{1}{\Delta t^*}\right)\Delta u
+\mathcal{N}(\mathbf{p})\!\left[u^n+\Delta u\right]
-\mathcal{N}(\mathbf{p}^*)\!\left[u^*+\Delta u\right].
\label{eq:increment_operator_difference}
\end{equation}
The time-discretized equation can then be recast as
\begin{equation}
\mathcal{O}^*\!\left[\Delta u^{n+1}\right]
=
S^{n+1}-\mathcal{O}_\delta\!\left[\Delta u^{n+1}\right]
\equiv S_{\mathrm{eff}}^{n+1}.
\label{eq:increment_recast}
\end{equation}
This is an exact identity for the time-discretized equation.
Parameter variation, time-step variation, and changes in the current field are incorporated into the effective source, while the canonical operator remains fixed.

With the learned canonical inverse $\mathcal{G}^*_N\approx(\mathcal{O}^*)^{-1}$, the target increment is predicted through the fixed-point iteration
\begin{equation}
\Delta u^{n+1,(m+1)}
=
\mathcal{G}^*_N\!\left[S_{\mathrm{eff}}^{n+1,(m)}\right],
\qquad
S_{\mathrm{eff}}^{n+1,(m)}
=
S^{n+1}
-\mathcal{O}_\delta\!\left[\Delta u^{n+1,(m)}\right],
\qquad m=0,1,2,\dots,
\label{eq:increment_iteration}
\end{equation}
where $n$ indexes the time step and $(m)$ denotes the recast iteration.
After convergence, the updated field is obtained as $u^{n+1}=u^n+\Delta u^{n+1}$.

\subsection{Geometry canonicalization}
\label{sec:method:geometry}

In the finite element method, it is standard practice to map irregular elements to a canonical reference element, where integrals are evaluated using the associated Jacobian metrics. Inspired by this idea, equation recast is extended to handle geometry variation by mapping different physical domains into a shared canonical domain. In the present study, we focus on two-dimensional simply connected domains, for which such mappings can be constructed. Let \(\boldsymbol{\xi} = (\xi_1,\xi_2)\) denote the canonical coordinates and \(\mathbf{x} = (x,y)\) the physical coordinates. The mapping from canonical to physical coordinates is written as
\begin{equation}
    \mathbf{x} = \phi(\boldsymbol{\xi}),
\end{equation}
with Jacobian
\begin{equation}
    \mathrm{K}(\boldsymbol{\xi}) = \frac{\partial \mathbf{x}}{\partial \boldsymbol{\xi}}.
\end{equation}
Under this transformation, geometric variation is represented explicitly through the Jacobian and the associated metric terms. In the physical domain, the governing equation is written abstractly as
\begin{equation}
    \mathcal{O}_\Omega(\mathbf{p})[u] = S.
\end{equation}
Here \(\Omega\) denotes the physical domain. After mapping to the canonical domain, geometry dependence is represented analytically through the Jacobian metrics \(\mathrm{K}\) and the transformed equation becomes
\begin{equation}
    \tilde{\mathcal{O}}(\mathbf{p}, \mathrm{K})[\tilde{u}] = \tilde{S},
\end{equation}
where the geometry dependence is incorporated analytically through the Jacobian \(\mathrm{K}\). Choosing a canonical parameter set \(\mathbf{p}^*\) and a reference geometry with Jacobian \(\mathrm{K}^*\), the canonical operator is defined as
\begin{equation}
    \tilde{\mathcal{O}}^* \equiv \tilde{\mathcal{O}}(\mathbf{p}^*, \mathrm{K}^*).
\end{equation}
For a new geometry, the recast form of the PDE becomes
\begin{equation}
    \tilde{\mathcal{O}}^*[\tilde{u}]
=
\tilde{S} - \tilde{\mathcal{O}}_{\delta}(\delta \mathbf{p}, \delta \mathrm{K})[\tilde{u}],
\end{equation}
where \(\delta\mathrm{K}=\mathrm{K}-\mathrm{K}^*\), and \(\tilde{\mathcal{O}}_{\delta}\) collects both parameter- and geometry-induced variations. In this formulation, geometry variation is treated in the same framework as parameter variation, with all geometric effects entering analytically through the Jacobian.
The learned canonical-domain operator therefore acts on a common representation for all included geometries.
For each geometry, the mapping and its Jacobian are computed once and reused.

\subsection{Fixed-point consistency and convergence behavior}
\label{sec:method:fixedpoint}

The equation recast formulation induces a fixed-point iteration on the solution field. Here, we return to the general formulation in Section~\ref{sec:method:recast} to examine its consistency and convergence behavior.
Define the iteration mapping
\begin{equation}
\mathcal{T}[u] \;=\; \mathcal{G}^*_N\!\left[S - \mathcal{O}_\delta(\delta \mathbf{p})[u]\right],
\label{eq:iteration_mapping}
\end{equation}
whose fixed points $u^\dagger$ satisfy $\mathcal{T}[u^\dagger] = u^\dagger$. If the learned canonical inverse $\mathcal{G}^*_N$ recovers $(\mathcal{O}^*)^{-1}$ exactly, 
applying $\mathcal{O}^*$ to the fixed-point equation and substituting the definition of $\mathcal{O}_\delta$ gives
\begin{equation}
\mathcal{O}(\mathbf{p})[u^\dagger] \;=\; S,
\end{equation}
so that any converged iterate is a solution of the original PDE at parameter $\mathbf{p}$. The recast iteration is therefore consistent with the governing equation under exact operator inversion.

In practice, approximation error is unavoidable. For an arbitrary input field $y$ in the source space, define the inverse residual
\begin{equation}
r(y) \;\equiv\; \mathcal{O}^*\!\left[\mathcal{G}^*_N[y]\right] - y,
\label{eq:inverse_residual}
\end{equation}
which vanishes when $\mathcal{G}^*_N$ is exact. A fixed point of the approximate iteration then satisfies
\begin{equation}
\mathcal{O}(\mathbf{p})[u^\dagger] 
\;=\; S + r\!\left(S - \mathcal{O}_\delta(\delta \mathbf{p})[u^\dagger]\right),
\label{eq:fp_residual}
\end{equation}
so that the original PDE is satisfied up to a residual induced by the learned canonical inverse. The magnitude of this residual is controlled jointly by the inverse approximation error, through $r$, and by the effective source $S - \mathcal{O}_\delta(\delta \mathbf{p})[u^\dagger]$, which reflects the deviation from the canonical regime. 

A sufficient local condition for fixed-point convergence is that the iteration map
\(\mathcal{T}\) be contractive in the relevant solution-space metric.
In practice, the corresponding Lipschitz constants of a learned neural operator are not available a priori, so we do not use such a bound as a predictive criterion.
Instead, convergence is assessed directly from the observed iteration.
Standard relaxation or acceleration strategies, including under-relaxation, Anderson mixing~\cite{anderson1965iterative}, and Aitken's \(\Delta^2\) method~\cite{aitken1926bernoulli}, can be used to improve numerical stability or convergence rate without changing the target fixed point when the iteration converges.
When the iteration converges, it yields a self-consistent recast solution whose final accuracy also depends on the learned canonical inverse.
Slow convergence or divergence instead indicates a reduced convergence margin of the chosen recast continuation pathway.
This behavior is illustrated most clearly by the Helmholtz example, where proximity to resonance produces rapidly increasing iteration counts and eventual loss of convergence.

\section{Results}
\label{sec:results}

In this section, equation recast is evaluated across a sequence of PDE benchmarks designed to isolate its key properties, followed by a high-fidelity tokamak electron-temperature simulation application. Additional details on the problem configurations, numerical setups, and extra ablation tests are provided in the Appendices.

\subsection{Equation recast enables zero-shot extrapolation across multiple PDE parameters}
\label{sec:results:adr}
\begin{figure}[h]
    \centering
    \includegraphics[width=0.9\textwidth]{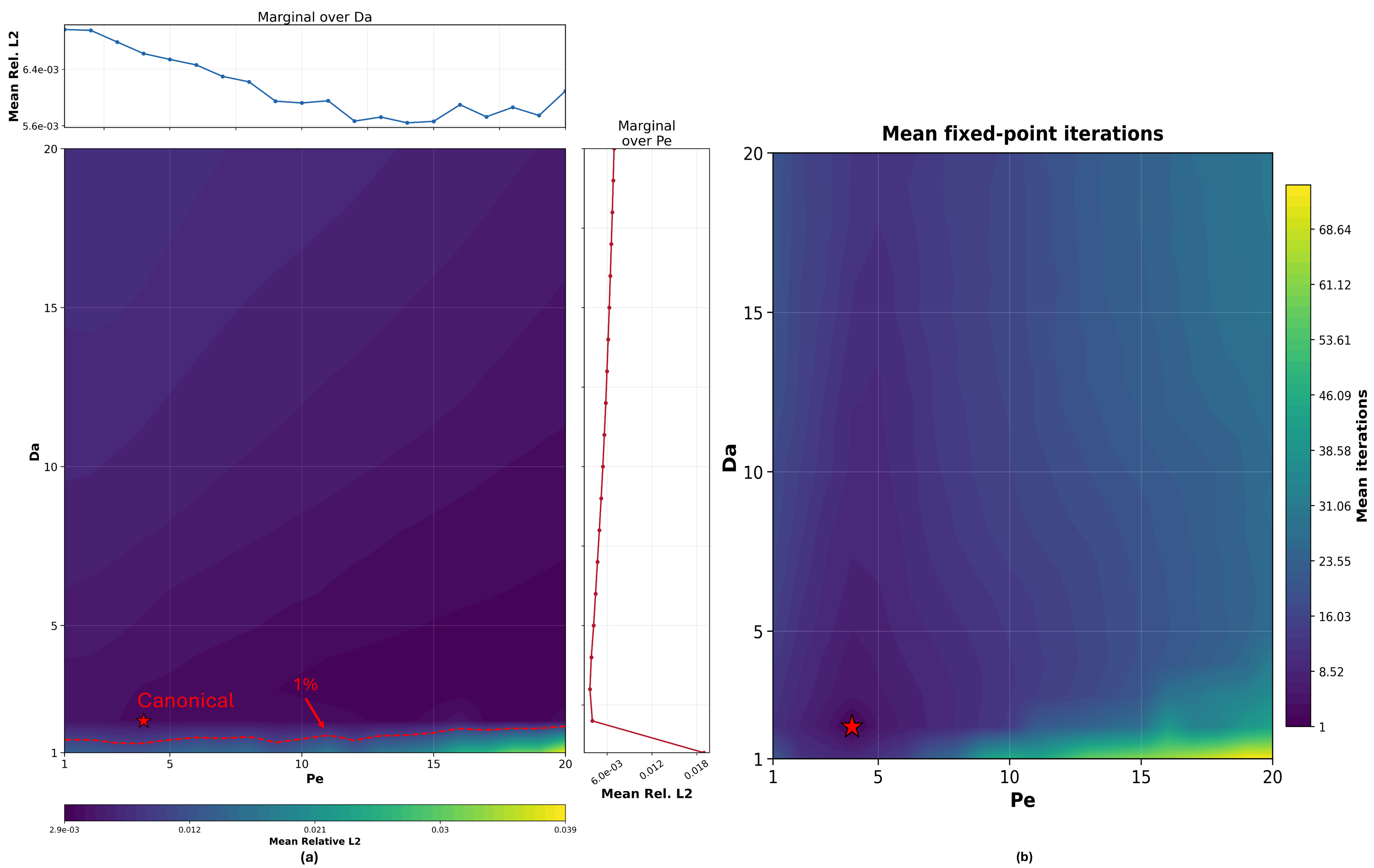}
    \caption{\textbf{Equation recast enables zero-shot parametric extrapolation from a single canonical operator.} \textbf{(a)} One-dimensional advection-diffusion-reaction (ADR) equation: averaged relative \(L_2\) error over the \((\mathrm{Pe},\mathrm{Da})\) parameter plane using a canonical neural operator trained only at \((\mathrm{Pe}^*,\mathrm{Da}^*)=(4,2)\) (red star). 
    The red dashed contour marks the \(1\%\) relative-error level, and the top and right panels show the marginal error profiles along \(\mathrm{Da}\) and \(\mathrm{Pe}\), respectively. 
    The extrapolation error is weakly sensitive to \(\mathrm{Pe}\) over the tested range, but becomes more sensitive to decreasing \(\mathrm{Da}\). 
    \textbf{(b)} Mean number of recast iterations over the same parameter domain.
    The iteration count increases most strongly in the high-\(\mathrm{Pe}\), low-\(\mathrm{Da}\) regime and follows the main degradation in prediction accuracy.
    }
    \label{fig:Figure2}
\end{figure}

In the equation recast framework, parameter variation is not learned explicitly, but is instead absorbed analytically into an effective source defined with respect to a fixed canonical operator.
We first examine whether a single canonical neural operator $\mathcal{G}^*_N$ can support zero-shot extrapolation across parameter variation using the one-dimensional advection diffusion reaction (ADR) equation
\begin{equation}
    -u''(x) + \mathrm{Pe}\,u'(x) + \mathrm{Da}\,u(x) = S(x),
\end{equation}
on a periodic domain \(x\in[0,1]\), where \(\mathrm{Pe}\) and \(\mathrm{Da}\) denote the P\'eclet and Damk\"ohler numbers. 
A canonical neural operator is trained only at the reference configuration \((\mathrm{Pe}^*,\mathrm{Da}^*)=(4,2)\), while parameter variation is incorporated analytically through the recast formulation (Table~\ref{tab:equation_recast_cases}). 
Figure~\ref{fig:Figure2}(a) shows the mean relative $L^2$ error across test samples over the parameter domain $(\mathrm{Pe},\mathrm{Da})\in[1,20]^2$. 
Despite training at a single parameter point, the model maintains low error over a broad region, demonstrating accurate zero-shot extrapolation across the two-dimensional parameter space. 
The largest degradation occurs in the high-$\mathrm{Pe}$, low-$\mathrm{Da}$ regime, where the derivative correction is consistent with reduced contractivity of the fixed-point update, as reflected by the increased iteration count in Figure~\ref{fig:Figure2}(b).

\begin{figure}[htbp]
    \centering
    \includegraphics[width=0.9\textwidth]{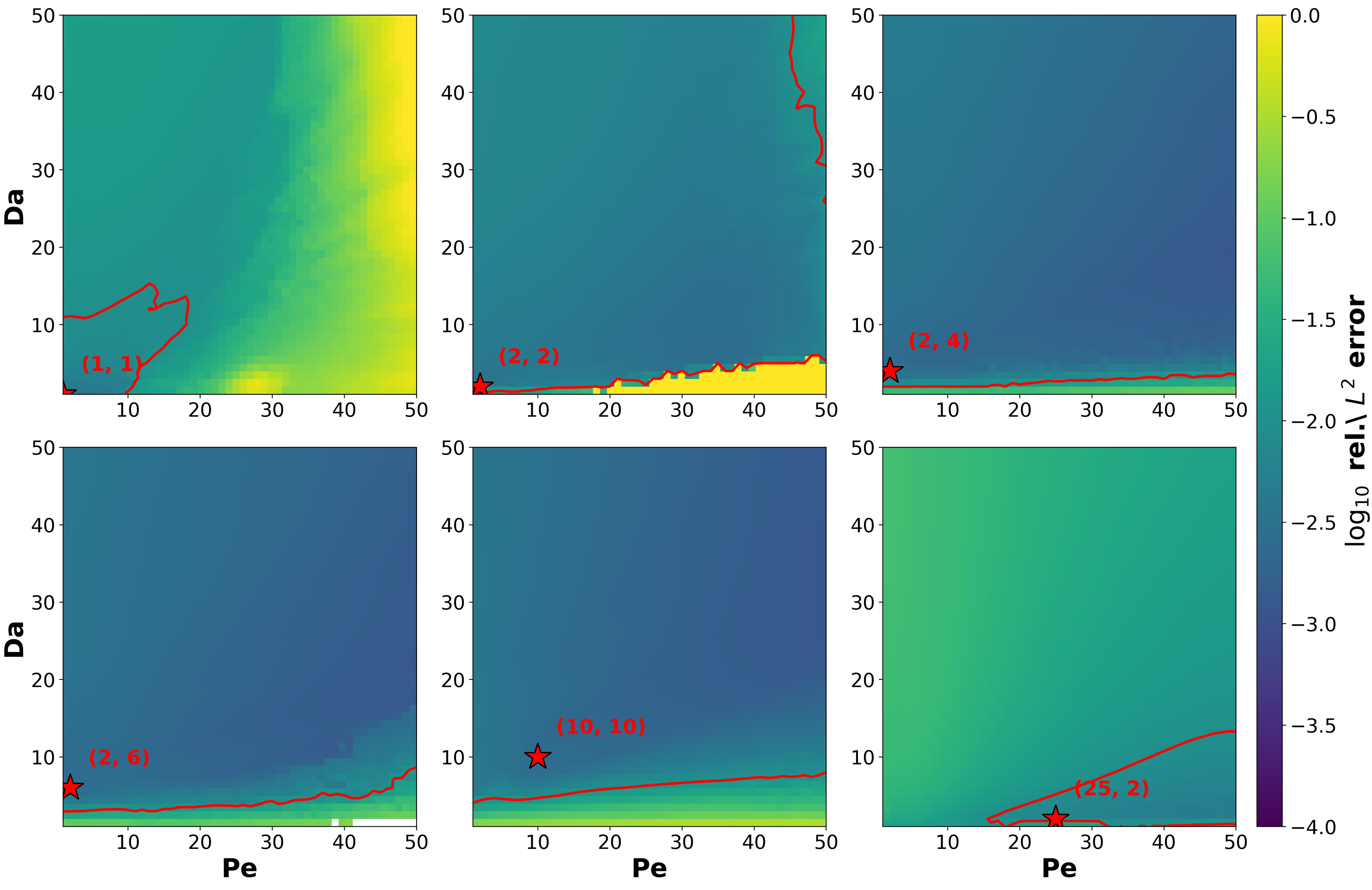}
    \caption{
    \textbf{Sensitivity of ADR extrapolation to the canonical configuration.}
    Mean relative \(L^2\) error (\(\log_{10}\) scale) over
    \((\mathrm{Pe},\mathrm{Da})\in[1,50]^2\)
    for six canonical configurations:
    top row, \((1,1)\), \((2,2)\), and \((2,4)\);
    bottom row, \((2,6)\), \((10,10)\), and \((25,2)\).
    Red contours indicate the \(1\%\) relative-error boundary and red stars mark the corresponding canonical configurations.
    Broad reusable regions are retained across the intermediate reference configurations, whereas the accessible domain contracts strongly for the low-\(\mathrm{Pe}\), low-\(\mathrm{Da}\) configuration \((1,1)\) and the advection-dominated configuration \((25,2)\).
    }
    \label{fig:Figure_s1}
\end{figure}

To assess sensitivity to the choice of canonical configuration, the ADR benchmark is repeated for six canonical points,
\begin{equation}
    (\mathrm{Pe}^*,\mathrm{Da}^*)
    \in
    \left\{
    (1,1),
    (2,2),
    (2,4),
    (2,6),
    (10,10),
    (25,2)
    \right\},
\end{equation}
over the extended parameter domain
\((\mathrm{Pe},\mathrm{Da})\in[1,50]^2\)
with unit spacing.
All canonical models use the same source distribution and training-set size and are evaluated on the same test distribution.
Figure~\ref{fig:Figure_s1} shows that the reusable region is relatively insensitive to the precise canonical location over a broad range of reference configurations.
The canonical configurations \((2,2)\), \((2,4)\), and \((2,6)\) retain approximately \(92\)--\(96\%\) of the extended parameter domain below \(1\%\) mean relative \(L^2\) error (red contours), with a similarly broad reusable region obtained for \((10,10)\).
In contrast, the reusable region contracts sharply for the low-\(\mathrm{Pe}\), low-\(\mathrm{Da}\) configuration \((1,1)\) and the advection-dominated configuration \((25,2)\), for which less than \(10\%\) of the scanned domain remains below the \(1\%\) error threshold.
These results indicate that the canonical point does not require fine tuning within a suitable operating regime, whereas extreme reference configurations can substantially restrict the accessible extrapolation domain.

\subsection{Equation recast integrates sparse heterogeneous data through effective source enrichment}
\label{sec:results:dr}
In high-fidelity,  multiphysics simulations, training data are often sparse and unevenly distributed across parameter space because each simulation is costly and only a limited number of operating conditions can be sampled.
This poses a challenge for parametric operator learning when some parameter regimes are weakly represented during training.

Equation recast provides a direct way to combine data generated at different parameter configurations within a single canonical representation. 
For a data pair \((S,u)\) generated at parameter \(\mathbf{p}\), with canonical parameter \(\mathbf{p}^*\), the corresponding effective source is
\begin{equation}
    S_{\mathrm{eff}} = S - \mathcal{O}_{\delta}(\delta \mathbf{p})[u].
\end{equation}
During training, \(S_{\mathrm{eff}}\) is used as the model input while the original high-fidelity solution \(u\) remains the supervision target.
Data collected away from the canonical parameter can therefore be recast and used jointly with canonical data to train a single inverse operator.
Importantly, a new parameter configuration does not merely map to a single point in the canonical representation, however, it generates a distribution of effective sources across various source realizations.
Heterogeneous data can therefore expand the coverage of the canonical effective-source space by populating regions that are weakly represented, or absent, in data generated at the canonical parameter alone.

This mechanism is demonstrated using the one-dimensional reaction-diffusion equation,
\begin{equation}
    -\frac{d^2 u(x)}{dx^2} + k\,u(x) = S(x), \qquad k>0,
\end{equation}
with canonical parameter \(k^*=0.785\). 
Two models are trained with the same total number of samples.
Model~1 uses only data generated at \(k^*\), whereas for Model~2 half of the canonical samples are replaced by data generated at \(k_{\mathrm{new}}=0.05\), which are recast into the same canonical effective-source representation before training.

\begin{figure}[htbp]
    \centering
    \includegraphics[width=0.9\textwidth]{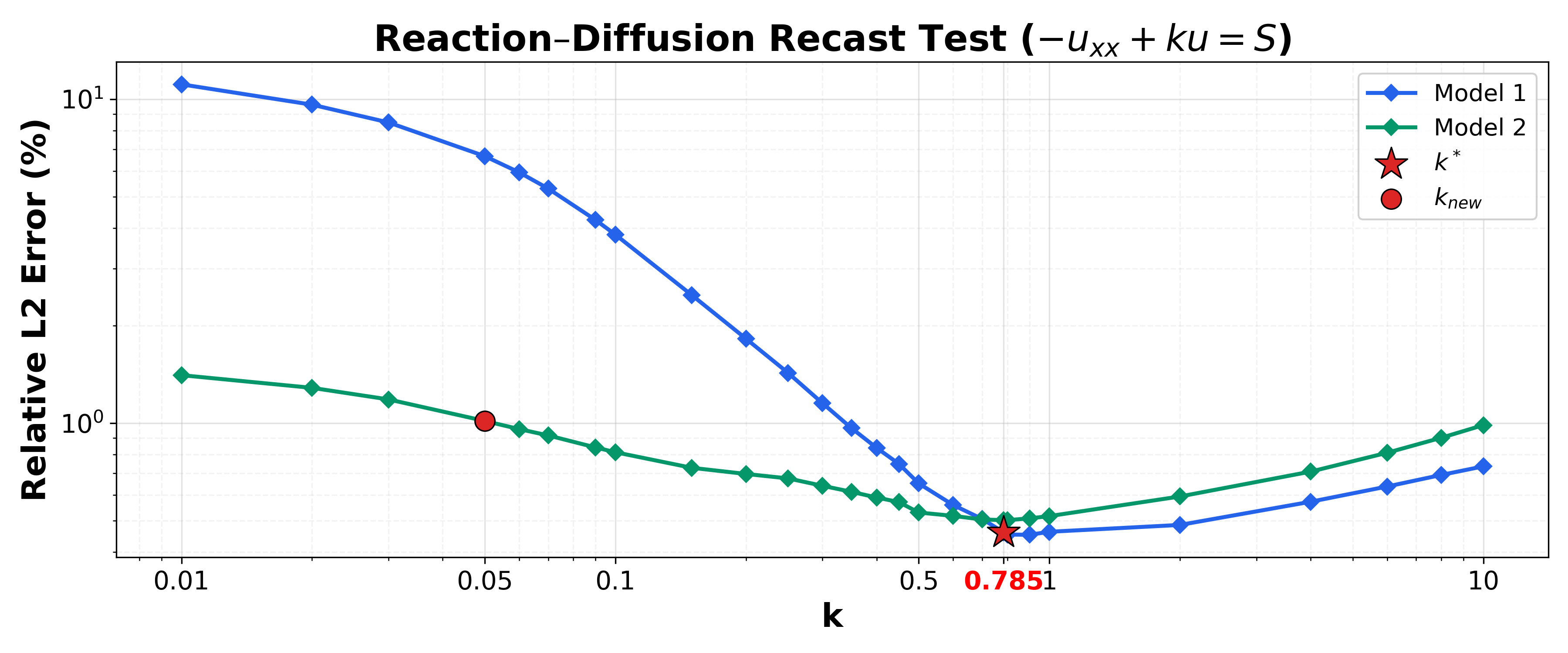}
    \caption{\textbf{Equation recast integrates heterogeneous data through effective-source enrichment.}
    In the one-dimensional reaction-diffusion equation example, mean relative \(L_2\) error is plotted as a function of \(k\) for two canonical neural operators trained with the same total number of samples. Model~1 uses only data generated at the canonical parameter \(k^*=0.785\) (red star), whereas for Model~2 half of the canonical samples are replaced by data generated at \(k_{\mathrm{new}}=0.05\) (red circle) and recast into the same canonical effective-source representation. Incorporating heterogeneous data substantially improves accuracy across the low-\(k\) regime, with the benefit extending well beyond the added parameter value, consistent with enrichment of a broader region of effective-source space. The modest loss of accuracy at large \(k\) reflects redistribution of the fixed training budget away from the canonical regime.}
    \label{fig:Figure3a}
\end{figure}

Figure~\ref{fig:Figure3a} shows that incorporating heterogeneous data in the low-\(k\) regime substantially changes the extrapolation behavior.
For \(k<k^*\), the canonical-only model exhibits rapidly increasing error as \(k\) moves away from the training configuration.
Including recast data from \(k_{\mathrm{new}}=0.05\) strongly suppresses this degradation, reducing the error by several-fold across the low-\(k\) regime and by nearly an order of magnitude at the most extrapolative values tested.
Notably, the improvement extends well beyond the added parameter value \(k_{\mathrm{new}}\), consistent with the heterogeneous samples enriching a broader region of the canonical effective-source space that is relevant to neighboring parameter values.
Near the canonical parameter, the two models attain comparable accuracy, whereas at large \(k\) Model~1 performs slightly better, consistent with redistributing a fixed training budget away from the canonical regime toward the low-\(k\) effective-source region.

These results demonstrate that equation recast can effectively utilize heterogeneous data collected away from the canonical configuration by transforming them into a shared canonical effective-source representation, allowing information from sparsely sampled parameter conditions to improve generalization beyond the conditions at which it was generated.
This source-space perspective further suggests a strategy for targeted data acquisition, in which new simulations are selected to enrich effective-source regions that are weakly represented by the existing training set.

\subsection{Equation recast provides an iteration-based indicator of prediction reliability}
\label{sec:results:helmholtz}
Parameter extrapolation can encounter singular parameter values at which the underlying inverse operator becomes ill-conditioned or unbounded.
Because equation recast performs extrapolation through a fixed-point iteration, loss of operator regularity can be reflected directly in its convergence behavior, providing a numerical indicator of when extrapolation from a given canonical operator is no longer reliable.

We illustrate this behavior using the one-dimensional Helmholtz equation,
\begin{equation}
    \frac{d^2u(x)}{dx^2}+k^2u(x)=-S(x),
\end{equation}
defined on \(x\in[0,L]\) with \(L=10\) and Dirichlet boundary conditions \(u(0)=u(L)=0\).
The resonance points are
\begin{equation}
    k_n=\frac{n\pi}{L}, \qquad n=1,2,3,\ldots,
\end{equation}
giving \(k_2=0.628\), \(k_3=0.942\), and \(k_4=1.257\).
The canonical parameter is chosen as \(k^*=0.785=\tfrac{1}{2}(k_2+k_3)\), between the second and third resonances.
To test whether additional heterogeneous data can extend the usable regime, a second model is trained by replacing part of the canonical training set with data generated at \(k_{\mathrm{new}}=1.099=\tfrac{1}{2}(k_3+k_4)\) and recast into the canonical representation.

\begin{figure}[htbp]
    \centering
    \includegraphics[width=0.9\textwidth]{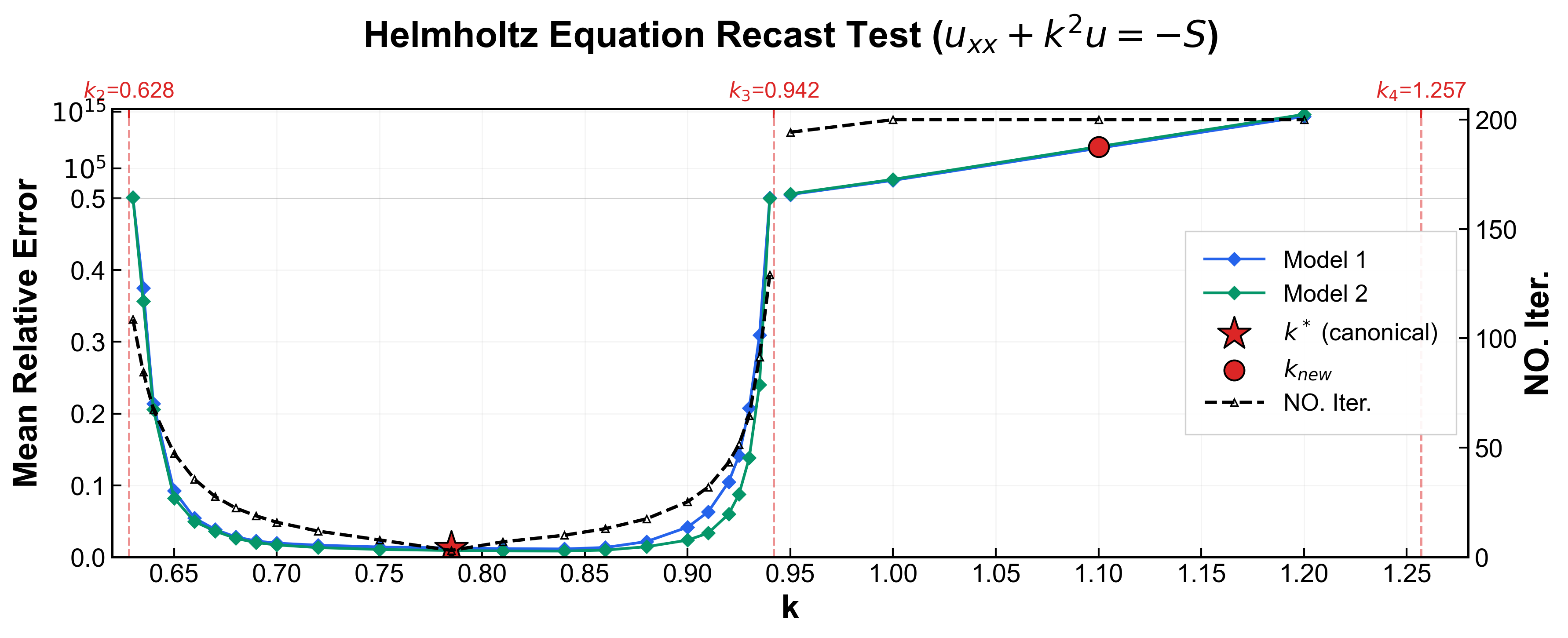}
    \caption{\textbf{Equation recast provides an iteration-based indicator of prediction reliability.}
    One-dimensional Helmholtz equation: mean relative \(L_2\) error for two fixed-budget training strategies, together with the number of recast iterations (black dashed curve, right axis). The canonical parameter is \(k^*=0.785\), and Model~2 additionally incorporates recast data generated at \(k_{\mathrm{new}}=1.099\) (red circle). Vertical dashed red lines mark the resonance points \(k_2\), \(k_3\), and \(k_4\). Within \(k\in(k_2,k_3)\), the error and iteration count both increase as either resonance is approached. Although the additional heterogeneous data reduce the pre-resonance error near \(k_3\), neither model recovers accurate extrapolation beyond \(k_3\): throughout the tested portion of \(k\in(k_3,k_4)\), the iteration reaches the prescribed limit and the prediction error increases by many orders of magnitude. The rapidly increasing iteration count and eventual loss of convergence therefore provide a direct diagnostic of diminishing convergence margin and failure of stable continuation from the chosen canonical operator.}
    \label{fig:Figure3b}
\end{figure}

Figure~\ref{fig:Figure3b} shows the mean relative \(L_2\) error together with the number of recast iterations.
Within the resonance interval containing the canonical point, \(k\in(k_2,k_3)\), the error is lowest near \(k^*\) and rises sharply as either resonance is approached.
The iteration count follows the same trend, increasing toward both singular boundaries.
Model~2 reduces the pre-resonance error as \(k\) approaches \(k_3\) from the canonical side, consistent with the effective-source enrichment observed above.
However, neither model recovers accurate extrapolation beyond \(k_3\).
Throughout the tested portion of \(k\in(k_3,k_4)\), the recast iteration reaches the prescribed iteration limit and the prediction error increases by many orders of magnitude despite the inclusion of training data at \(k_{\mathrm{new}}\).

This failure is governed by the operator structure rather than by insufficient training coverage.
At a Helmholtz resonance, the inverse operator is singular and its resolvent norm diverges.
As the target parameter approaches such a singularity, the recast fixed-point map progressively loses contractivity. 
Therefore, across the resonance, the iteration associated with the chosen canonical operator becomes non-contractive.
Additional heterogeneous data can improve the approximation of the canonical inverse over a broader effective-source distribution, but cannot remove the underlying operator singularity, and convergence is no longer guaranteed once the sufficient contraction condition is lost.
Accordingly, the iteration count rises sharply as the resonance is approached and reaches the prescribed maximum once convergence is lost, closely tracking the onset of large extrapolation error (Figure~\ref{fig:Figure3b}).
The convergence behavior of the recast iteration therefore provides an interpretable indicator of prediction reliability: an increasing iteration count signals a diminishing convergence margin, while failure to converge indicates loss of convergence margin along the chosen recast continuation pathway.

\subsection{Equation recast enables data-efficient parameter extrapolation in nonlinear Navier–Stokes equations}
\label{sec:results:ns}
We next evaluate equation recast in the nonlinear, two-dimensional Navier–Stokes equations, and compare it directly with neural operators that learn parameter dependence from distributed training data.
Consider the steady incompressible Navier-Stokes equation in vorticity form on the periodic domain
\(\mathbb{T}^2=[0,1]^2\),
\begin{equation}
    \mathbf{u}\cdot\nabla\omega(x,y)
    =
    \frac{1}{\mathrm{Re}}\Delta\omega(x,y)
    +
    S(x,y),
    \label{eq:ns_vorticity}
\end{equation}
where \(\omega\) is the vorticity field, \(\mathbf{u}\) is the velocity field, \(S\) is a steady forcing, and \(\mathrm{Re}\) is the Reynolds number.
Under zero-mean periodic conditions, the velocity can be recovered from the vorticity through the stream function, so the problem can be written in operator form as
\begin{equation}
    \mathcal{O}(\mathrm{Re})[\omega]=S.
\end{equation}
For the canonical Reynolds number \(\mathrm{Re}^*\), equation recast gives
\begin{equation}
    \mathcal{O}^*[\omega]
    =
    S+
    \left(
        \frac{1}{\mathrm{Re}}
        -
        \frac{1}{\mathrm{Re}^*}
    \right)\Delta\omega,
\end{equation}
so that Reynolds-number variation is absorbed into an effective-source correction while the inverse operator remains fixed.

The canonical neural operator \(\mathcal{G}_N^*\approx(\mathcal{O}^*)^{-1}\) is represented by a two-dimensional FNO and trained using \(200\) samples generated only at \(\mathrm{Re}^*=250\).
We compare it with two conditioned baselines built on the same FNO backbone and trained using the same \(200\) supervised samples distributed over \(\mathrm{Re}\in[200,300]\) a parametric FNO, for which the Reynolds number is supplied as an additional input channel, and a physics-informed neural operator (PINO), which additionally incorporates the governing-equation residual into the training objective.
The comparison therefore contrasts two uses of the same supervised data budget: concentrating the samples at a single canonical operator and handling parameter variation analytically during inference, or distributing the samples across parameter space and learning the parameter dependence directly with a conditioned neural operator.

\begin{figure}[htbp]
    \centering
    \includegraphics[width=0.90\textwidth]{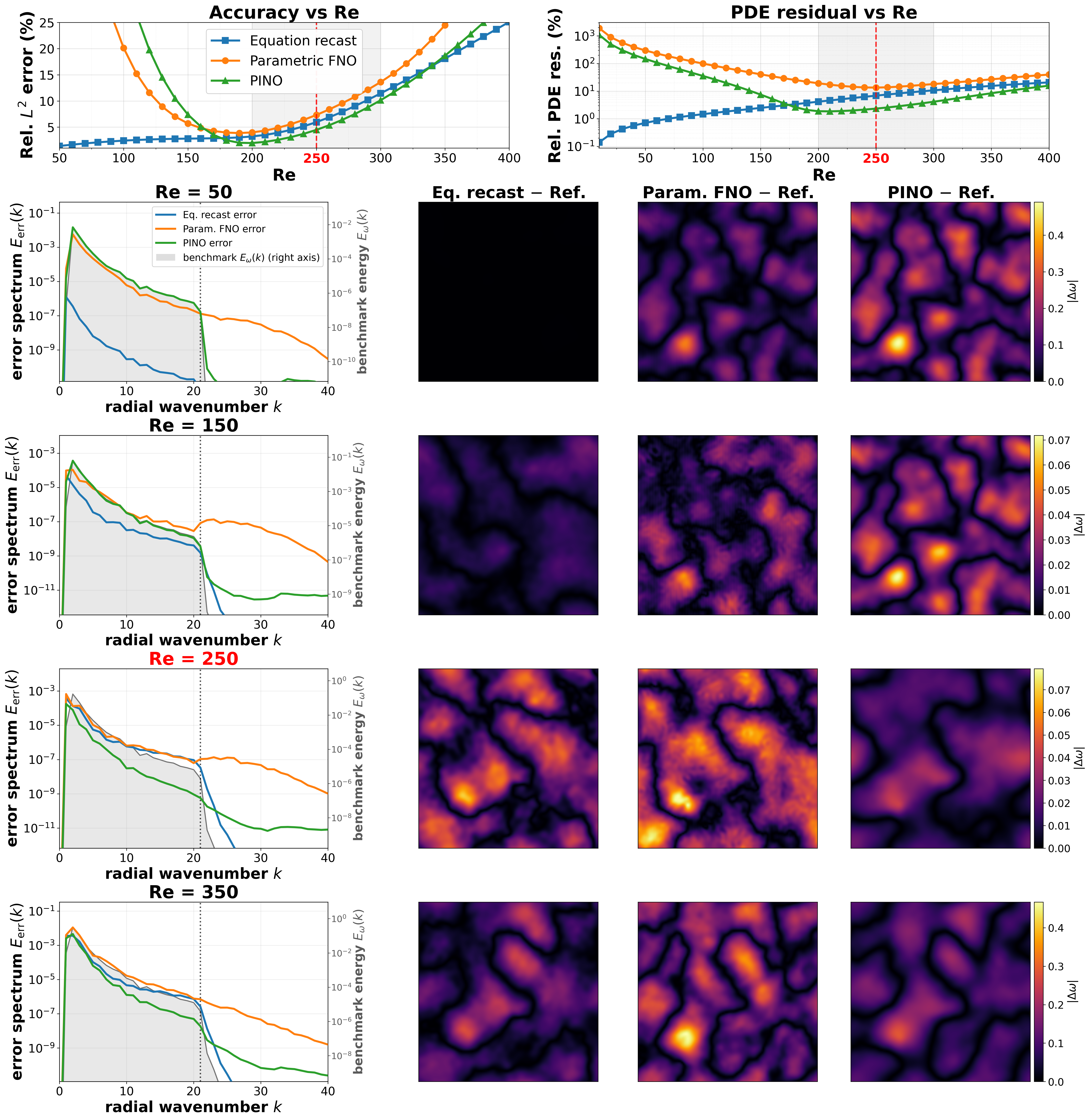}
    \caption{
    \textbf{Data-efficient extrapolation and consistency assessment in nonlinear Navier-Stokes dynamics.}
    \textbf{Top:} Mean relative \(L^2\) error (left) and normalized PDE residual (right) as functions of Reynolds number, averaged over \(20\) test forcing realizations.
    Equation recast uses a canonical FNO trained with \(200\) samples at the single reference Reynolds number \(\mathrm{Re}^*=250\) (red dashed line).
    The parametric FNO and PINO use the same FNO backbone and the same \(200\) supervised samples distributed over \(\mathrm{Re}\in[200,300]\) (grey shaded region); PINO additionally incorporates a PDE-residual term during training.
    The conditioned models retain high accuracy within their sampled parameter interval but degrade rapidly outside it, whereas equation recast remains accurate over a substantially broader range, particularly toward lower Reynolds numbers.
    \textbf{Bottom:} Spectral and spatial decomposition of the prediction error at \(\mathrm{Re}=50\), \(150\), \(250\), and \(350\).
    The left column shows the radial error spectrum \(E_{\mathrm{err}}(k)\) for equation recast, the parametric FNO, and PINO, together with the benchmark vorticity energy spectrum \(E_{\omega}(k)\) on the secondary axis.
    The vertical dotted line marks \(k=21\), the upper wavenumber of the band-limited forcing distribution used during data generation.
    The remaining columns show the 2D pointwise error fields \(\left|\omega_{\mathrm{pred}}-\omega_{\mathrm{ref}}\right|\) for equation recast, the parametric FNO, and PINO, using a common colour scale within each Reynolds-number row.
    At low Reynolds number, equation recast suppresses errors across the energy-containing modes and produces only weak spatial differences.
    At larger Reynolds number, its error increases predominantly in the resolved large-scale modes.
    }
    \label{fig:Figure4}
\end{figure}

Figure~\ref{fig:Figure4} shows the extrapolation performance over \(\mathrm{Re}\in[50,400]\).
Within the parameter interval sampled by the conditioned baselines, all three approaches retain relatively low error, with PINO benefiting from its additional physics-informed training objective.
Outside this interval, the conditioned models degrade rapidly, whereas equation recast maintains accurate predictions over a substantially broader range, particularly toward Reynolds numbers below the canonical value.
The recast error also increases progressively toward larger Reynolds numbers, showing that analytical recasting extends the usable range of the canonical inverse but does not remove the approximation limits of the learned operator itself.
Figure~\ref{fig:Figure4} also reports the normalized residual of the target governing equation.
Across the Reynolds-number scan, its overall trend is consistent with the degradation in reference-based prediction accuracy, while PINO attains particularly low residuals within and near its training regime, as expected from the residual term used during training.
The lower panels of Figure~\ref{fig:Figure4} further localize the extrapolation error in spectral and physical space.
At low Reynolds number, equation recast strongly suppresses errors across the energy-containing modes and produces only weak spatial differences from the reference solution.
As \(\mathrm{Re}\) increases, the error grows predominantly within the resolved large-scale modes that carry most of the vorticity energy, accompanied by coherent differences in the predicted flow structure.
The degradation is therefore associated with loss of accuracy in the dominant solution structure rather than the emergence of an unresolved small-scale error.

\begin{figure}[htbp]
    \centering
    \includegraphics[width=0.98\textwidth]{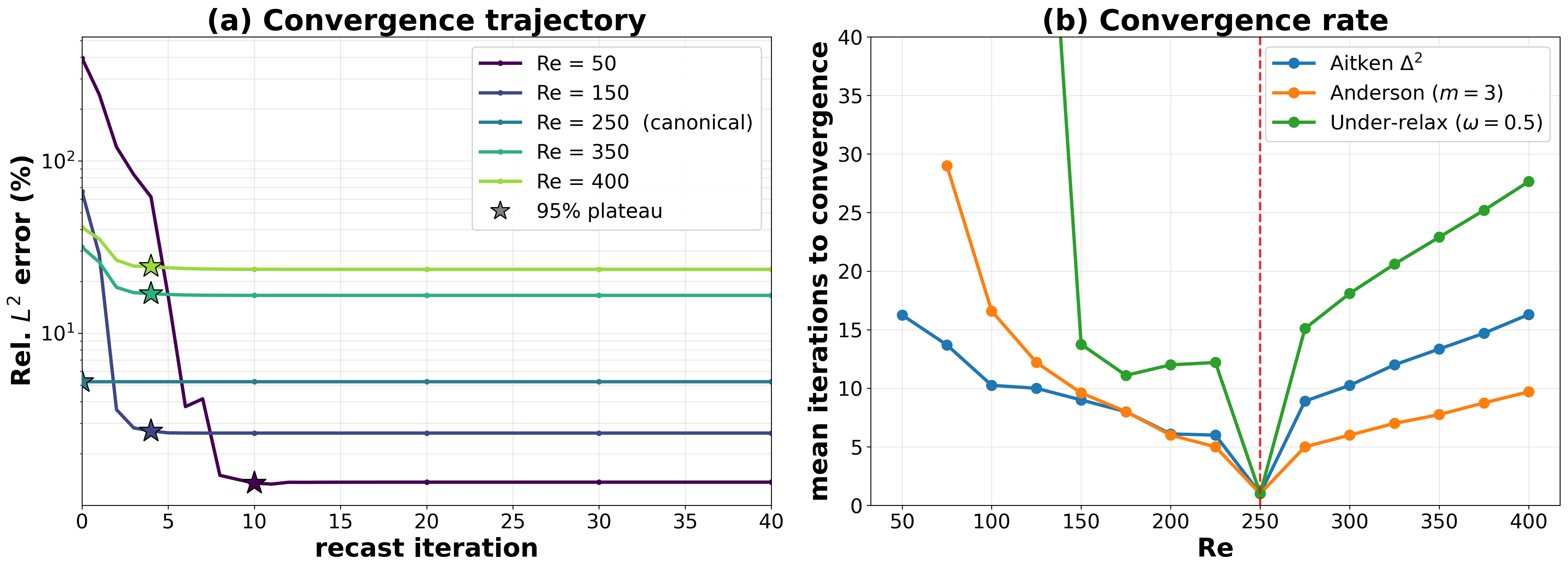}
    \caption{
    \textbf{Convergence behavior of the recast iteration for the Navier--Stokes benchmark.}
    \textbf{(a)} Relative \(L^2\) error of the recast iterate as a function of the iteration number at \(\mathrm{Re}=50\), \(150\), \(250\), \(350\), and \(400\) for the canonical FNO with Aitken \(\Delta^2\) relaxation, run for a fixed budget of \(80\) iterations without a stopping criterion. Curves are medians over \(20\) test sources processed as one batch.
    Stars mark the first iteration at which the error is within \(5\%\) of its plateau value, defined as the minimum error reached within the \(80\) iterations.
    \textbf{(b)} Mean iterations to convergence as a function of Reynolds number for Aitken \(\Delta^2\) relaxation, Anderson mixing (\(m=3\)), and fixed under-relaxation (\(\alpha=0.5\)), with each source iterated individually.
    Convergence is declared when the relative update of the iterate falls below \(10^{-5}\), with a maximum of \(200\) iterations.
    Results are averaged over \(20\) test sources at each Reynolds number.
    Non-converged cases are excluded from the iteration-count averages, and the vertical axis is truncated at \(40\) iterations.
    The canonical configuration is \(\mathrm{Re}^*=250\).
    }
    \label{fig:ns-convergence}
\end{figure}

Figure~\ref{fig:ns-convergence} characterizes the convergence of the recast fixed-point iteration.
In Fig.~\ref{fig:ns-convergence}(a), the iteration is advanced with Aitken's \(\Delta^2\) relaxation for a fixed budget of \(80\) iterations without any stopping criterion, so that the complete error trajectory is recorded.
The \(20\) test sources are processed as a single batch on the GPU, the relative \(L^2\) error of every iterate is measured against the reference solution at the target Reynolds number, and the median over sources is reported.
The plateau of a trajectory is defined as the minimum error reached within the \(80\) iterations, and the marked iteration is the first at which the error falls within \(5\%\) of this plateau.
In Fig.~\ref{fig:ns-convergence}(b), each source is instead iterated individually from the same initial iterate until the relative update \(\|\omega^{(m+1)}-\omega^{(m)}\|_2/\|\omega^{(m)}\|_2\) falls below \(10^{-5}\), with a maximum of \(200\) iterations, for the three relaxation schemes compared.
The iteration count is smallest near the canonical Reynolds number \(\mathrm{Re}^*=250\), and increases toward the edges of the Reynolds-number scan as a larger iterative correction is required, characterizing the numerical effort required to reach the recast fixed point.

\begin{figure}[htbp]
    \centering
    \includegraphics[width=0.94\textwidth]{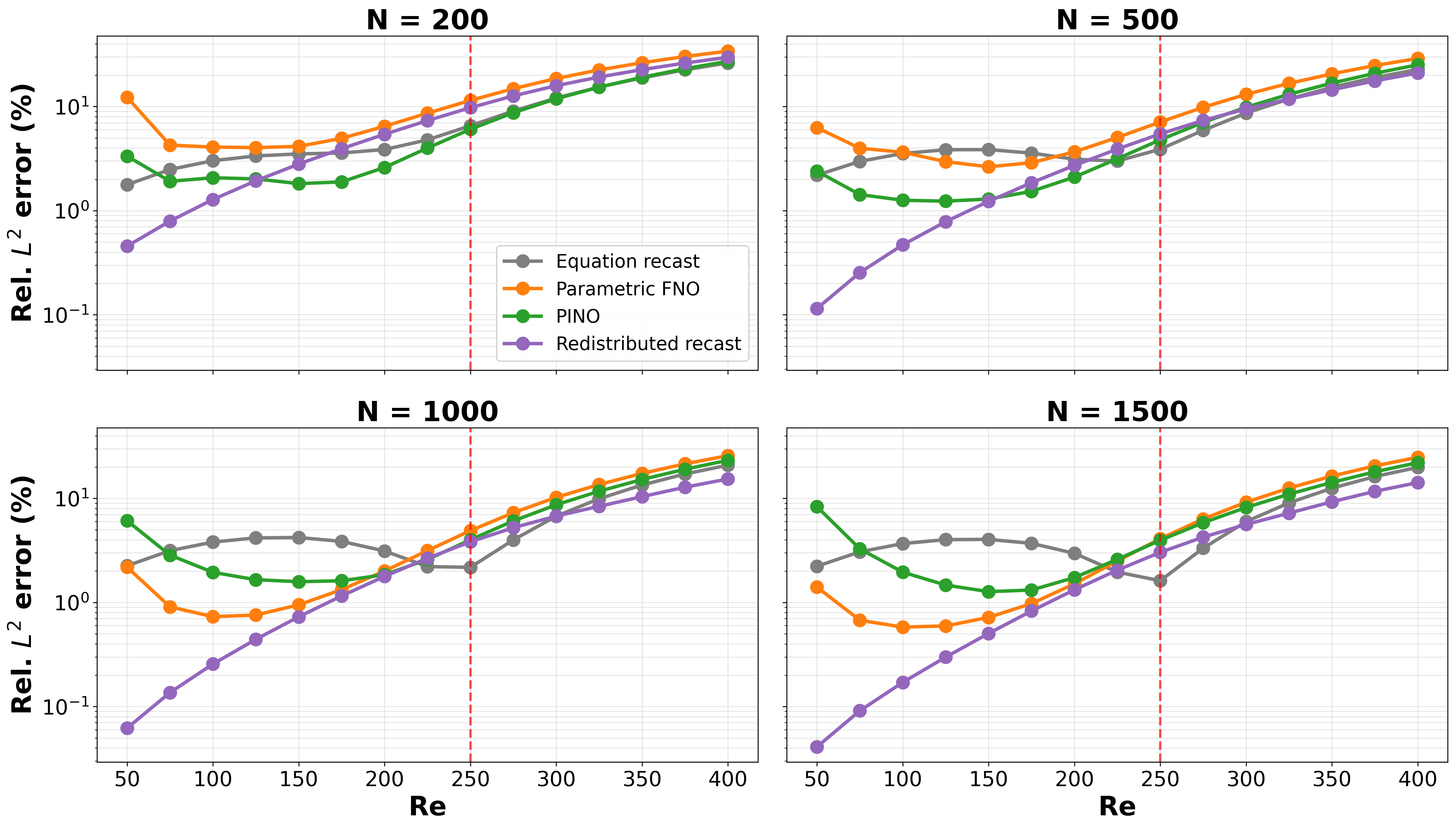}
    \caption{
    \textbf{Effect of training-data budget on Navier--Stokes generalization.}
    Relative \(L^2\) error as a function of Reynolds number for
    \(N=200\), \(500\), \(1000\), and \(1500\) supervised training samples.
    The four model families are the canonical equation-recast model, the full-range parametric FNO, PINO, and the redistributed equation-recast model trained with equal numbers of recast samples from
    \(\mathrm{Re}=50\), \(100\), \(200\), \(300\), and \(400\).
    The red dashed line marks the canonical Reynolds number
    \(\mathrm{Re}^*=250\).
    }
    \label{fig:si-budget-vs-Re}
\end{figure}

The role of training coverage is examined further through ablations for different dataset size $N$:
\[
    N\in\{200,500,1000,1500\}.
\]
In addition to the canonical and parametric training strategies, a redistributed equation-recast model is also included, in which heterogeneous samples generated at five discrete Reynolds numbers are transformed into the same canonical representation.
Three datasets containing up to \(1500\) samples are generated using the same reference solver.
The canonical dataset contains samples generated only at \(\mathrm{Re}^*=250\).
The full-range parametric dataset samples Reynolds numbers uniformly over \(\mathrm{Re}\in[50,400]\), covering the complete test interval.
The redistributed dataset contains equal-size blocks generated at
\[
    \mathrm{Re}\in\{50,100,200,300,400\}.
\]
For a sample \((S_i,\omega_i,\mathrm{Re}_i)\) from the redistributed dataset, the canonical training input is constructed as
\begin{equation}
    S_{\mathrm{eff},i}
    =
    S_i
    +
    \left(
        \frac{1}{\mathrm{Re}_i}
        -
        \frac{1}{\mathrm{Re}^*}
    \right)
    \Delta\omega_i,
    \label{eq:ns-redistributed-recast}
\end{equation}
while the corresponding reference solution \(\omega_i\) is retained as the supervision target.
At each total budget \(N\), the redistributed model uses \(N/5\) samples from each Reynolds-number block.
Four models are compared at every data budget, including the equation-recast model, the full-range parametric FNO, the PINO, and the redistributed equation-recast model.
All models use the same FNO backbone.
The PINO uses the same supervised training samples as the parametric FNO together with a PDE-residual contribution to the training objective with weight \(\lambda=0.5\).
Evaluation is performed using the same \(20\) held-out samples over \(\mathrm{Re}\in[50,400]\) with spacing \(25\).
Figure~\ref{fig:si-budget-vs-Re} shows the relative \(L^2\) error as a function of Reynolds number at each data budget.
At the smallest budget, the canonical recast and PINO provide the lowest prediction errors over most of the scan, whereas distributing only \(200\) samples over the full Reynolds-number interval leads to substantially larger error for the parametric FNO.
With increasing data availability, all model families improve, but the redistributed recast obtains the largest gain and progressively reduces the extrapolation error over most of the Reynolds-number range.
This result extends the heterogeneous-data mechanism demonstrated by the one-dimensional reaction--diffusion benchmark to a nonlinear two-dimensional PDE.

Together, these results demonstrate data-efficient parameter extrapolation in a nonlinear multiscale PDE using a neural operator trained at a single canonical parameter.

\subsection{Equation recast unifies high-fidelity multi-device Tokamak data in a shared canonical representation}
\label{sec:results:tokamak}
High-fidelity nonlinear magnetohydrodynamic (MHD) simulations are computationally intensive.
For large-scale simulation codes such as M3D-C1~\cite{doecode_12573}, resolving only milliseconds of plasma evolution can require weeks to months of computation on high-performance computing systems.
As a result, available high-fidelity simulation datasets are necessarily sparse.
Only a limited set of plasma conditions can be explored for any individual device, and the available data are also fragmented across machines with substantially different geometries.
This combination of high computational cost, sparse parameter coverage, and device-specific geometry makes it difficult to construct broadly reusable surrogate models using conventional parametric learning alone.
Here we examine whether equation recast can instead organize these heterogeneous simulation data into a single canonical learning problem.
Inspired by the standard finite-element paradigm~\cite{hughes2000fem}, irregular physical domains are first mapped to a shared canonical domain, where geometry enters the transformed governing equation analytically through Jacobian-dependent coefficients.
Equation recast then absorbs both plasma-coefficient and geometry-induced operator variations into the canonical representation, allowing data from different devices and operating conditions to contribute jointly to a single neural operator.

This idea is demonstrated using the electron temperature equation of high-fidelity M3D-C1 Tokamak simulations with realistic multiphysics setting relevant to thermal quench and massive gas injection~\cite{hollmann2015status,clauser2021modeling}.
M3D-C1 advances the extended-MHD system with an operator-split scheme, so that the electron temperature \(T_e\) is advanced by its own equation while the remaining fields enter through coefficients and source terms.
For the zero-flow configurations considered here, this equation reads
\begin{equation}
    \frac{\partial T_e}{\partial t}
    + \frac{\sigma_e}{n_e}\,T_e
    - \frac{\gamma-1}{n_e}\,\nabla\cdot\!\left(\kappa\nabla T_e\right)
    = Q_{\mathrm{tot}},
    \qquad
    Q_{\mathrm{tot}} = \frac{\gamma-1}{n_e}\left(\eta J^2 + Q_e + S_e\right),
    \label{eq:tokamak_physical}
\end{equation}
where \(n_e\) is the electron density, \(\sigma_e\) the ionization term, \(\gamma\) the adiabatic index, \(\kappa\) the isotropic thermal conductivity, which represents the dominant perpendicular transport, \(\eta J^2\) the ohmic heating, \(Q_e\) the radiated, ionization, and collisional-exchange power, and \(S_e\) an externally prescribed source.
The underlying simulations contain coupled plasma physics, whose effects on the electron-energy equation enter through the time- and space-dependent coefficient fields \(\mathbf{p}=(n_e,\sigma_e,\kappa)\) and the total heating and cooling source \(Q_{\mathrm{tot}}\).
The present benchmark models the external source, ionization, loss, and thermal-conduction contributions to the electron-temperature evolution in axisymmetric geometries rather than the full MHD state, and advective transport is not included.
To unify data from different devices in a common representation, each two-dimensional physical domain \(\Omega_g\) is mapped harmonically to a unit-disk canonical domain with coordinates \(\boldsymbol{\xi}\), \(\mathbf{x}=\phi(\boldsymbol{\xi})\), with Jacobian \(\mathrm{K}=\partial\mathbf{x}/\partial\boldsymbol{\xi}\) and \(|\mathrm{K}|=\det\mathrm{K}\).
Gradients transform as \(\nabla=\mathrm{K}^{-\mathsf{T}}\nabla_{\boldsymbol{\xi}}\) and the divergence carries the Jacobian determinant, so that the diffusion term becomes
\begin{equation}
    \mathcal{L}_{\kappa,\mathrm{K}}\!\left[\widetilde{T}_e\right]
    \equiv
    \frac{1}{|\mathrm{K}|}\,
    \nabla_{\boldsymbol{\xi}}\cdot
    \left(
        |\mathrm{K}|\,\tilde{\kappa}\,
        \mathrm{K}^{-1}\mathrm{K}^{-\mathsf{T}}
        \nabla_{\boldsymbol{\xi}}\widetilde{T}_e
    \right),
    \label{eq:tokamak_conduction_operator}
\end{equation}
where a tilde denotes a field expressed in canonical coordinates.
The canonical-domain form of Eq.~\eqref{eq:tokamak_physical} is then
\begin{equation}
    \frac{\partial \widetilde{T}_e}{\partial t}
    + \widetilde{\mathcal{N}}(\tilde{\mathbf{p}},\mathrm{K})\!\left[\widetilde{T}_e\right]
    = \widetilde{Q}_{\mathrm{tot}},
    \qquad
    \widetilde{\mathcal{N}}(\tilde{\mathbf{p}},\mathrm{K})\!\left[\widetilde{T}_e\right]
    \equiv
    \frac{\tilde{\sigma}_e}{\tilde{n}_e}\widetilde{T}_e
    - \frac{\gamma-1}{\tilde{n}_e}\,
    \mathcal{L}_{\kappa,\mathrm{K}}\!\left[\widetilde{T}_e\right],
    \label{eq:tokamak_canonical_domain}
\end{equation}
where all geometry dependence is carried by the Jacobian-related metric terms.
Following the increment formulation, backward-Euler discretization over a step \(\Delta t^{n+1}\) gives the time-advancement operator for the increment \(\Delta\widetilde{T}_e^{\,n+1}\equiv\widetilde{T}_e^{\,n+1}-\widetilde{T}_e^{\,n}\),
\begin{equation}
    \widetilde{\mathcal{O}}_{\mathrm{inc}}\!\left(\tilde{\mathbf{p}},\mathrm{K},\Delta t^{n+1};\widetilde{T}_e^{\,n}\right)\!\left[\Delta\widetilde{T}_e^{\,n+1}\right]
    \equiv
    \frac{\Delta\widetilde{T}_e^{\,n+1}}{\Delta t^{n+1}}
    + \widetilde{\mathcal{N}}(\tilde{\mathbf{p}},\mathrm{K})\!\left[\widetilde{T}_e^{\,n}+\Delta\widetilde{T}_e^{\,n+1}\right]
    = \widetilde{Q}_{\mathrm{tot}}^{\,n+1},
    \label{eq:tokamak_increment_operator}
\end{equation}
which depends on the coefficient fields, the geometry, the time step, and the current temperature field of the case.
The canonical operator is defined at a fixed reference state.
The canonical geometry is Alcator C-Mod with Jacobian \(\mathrm{K}^*\), and the reference coefficient fields \(\tilde{\mathbf{p}}^*=(\tilde{n}_e^*,\tilde{\sigma}_e^*,\tilde{\kappa}^*)\), the reference temperature field \(\widetilde{T}_e^{*}\), and the canonical step \(\Delta t^*\) are taken from one representative frame of a selected Alcator C-Mod simulation,
\begin{equation}
    \widetilde{\mathcal{O}}^{*}\!\left[\Delta\widetilde{T}_e\right]
    \equiv
    \widetilde{\mathcal{O}}_{\mathrm{inc}}\!\left(\tilde{\mathbf{p}}^*,\mathrm{K}^*,\Delta t^*;\widetilde{T}_e^{*}\right)\!\left[\Delta\widetilde{T}_e\right]
    =
    \frac{\Delta\widetilde{T}_e}{\Delta t^*}
    + \widetilde{\mathcal{N}}(\tilde{\mathbf{p}}^*,\mathrm{K}^*)\!\left[\widetilde{T}_e^{*}+\Delta\widetilde{T}_e\right].
    \label{eq:tokamak_onestep_operator}
\end{equation}
Equation recast then gives
\begin{equation}
    \widetilde{\mathcal{O}}^{*}\!\left[\Delta\widetilde{T}_e^{\,n+1}\right]
    =
    \widetilde{Q}_{\mathrm{tot}}^{\,n+1}
    - \widetilde{\mathcal{O}}_{\delta}\!\left[\Delta\widetilde{T}_e^{\,n+1}\right]
    \equiv
    \widetilde{S}_{\mathrm{eff}}^{\,n+1},
    \qquad
    \widetilde{\mathcal{O}}_{\delta}
    \equiv
    \widetilde{\mathcal{O}}_{\mathrm{inc}}\!\left(\tilde{\mathbf{p}},\mathrm{K},\Delta t^{n+1};\widetilde{T}_e^{\,n}\right)
    - \widetilde{\mathcal{O}}^{*},
    \label{eq:tokamak_recast}
\end{equation}
so that the plasma-coefficient variation \(\delta\tilde{\mathbf{p}}\), the geometry variation \(\delta\mathrm{K}\), and the deviation of the current temperature field from the reference field all enter through the effective source while the canonical operator remains fixed.
The learned neural operator \(\mathcal{G}_N^*\approx(\widetilde{\mathcal{O}}^{*})^{-1}\) predicts the temperature increment in the canonical domain.
\begin{equation}
    \Delta\widetilde{T}_e^{\,n+1}
    =
    \mathcal{G}_N^*\!\left[\widetilde{T}_e^{\,n},\widetilde{S}_{\mathrm{eff}}^{\,n+1}\right],
    \label{eq:tokamak_one_step}
\end{equation}
More details of the training pairs are given in Appendix~\ref{app:tokamak}. 

\begin{figure}[htbp]
    \centering
    \includegraphics[width=1.0\textwidth]{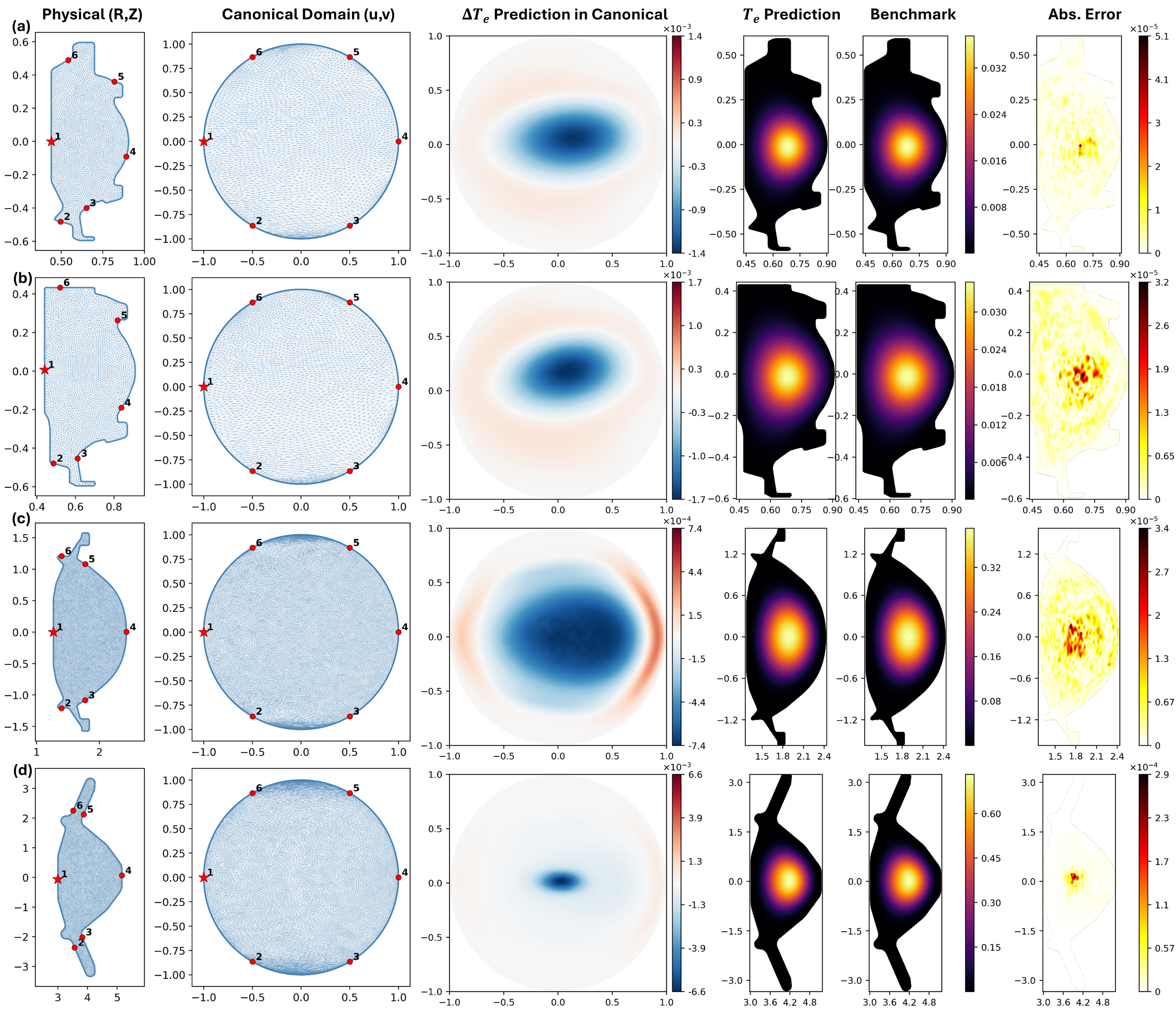}
    \caption{\textbf{Unification of parameter and geometry variation through canonical-domain equation recast.} Results for four Tokamak device geometries: (a) Alcator C-Mod, (b) C-Mod\_FD, (c) SPARC, and (d) ARC\_V2A. From left to right, the columns show the physical-domain mesh, the mapped canonical-domain mesh, the predicted electron-temperature increment \(\Delta T_e\) in the canonical domain, the mapped-back \(T_e\) prediction in the physical domain, the corresponding high-fidelity M3D-C1 benchmark, and the pointwise absolute error in the physical domain. All contours are in M3D-C1 internal unit. Despite substantial variation in device geometry, a single jointly trained canonical-domain neural operator produces accurate predictions across all configurations after Jacobian-aware equation recast without device-specific retraining. Geometry-induced variation is absorbed analytically through the mapping Jacobian, so that both parameter and geometry effects are treated within the same recast formulation.}
    \label{fig:Figure5}
\end{figure}

In this Tokamak demonstration, all device geometries are included during training, and the purpose is therefore to test whether geometry variations can be treated algebraically in the same recast form as parameter variation through the Jacobian-dependent metric terms. 
In this way, data produced by physically different devices contribute to a common learning problem rather than to device-specific surrogate models.
Figure~\ref{fig:Figure5} shows representative held-out predictions using the same jointly trained LocalNO~\cite{liuSchiaffini2024localizedNeuralOperators} across all four geometries, including Alcator C-Mod~\cite{clauser2024cmod}, C-Mod\_FD (Alcator C-Mod with a flat divertor)~\cite{labombard2008upperdiv}, SPARC~\cite{clauser2024sparc}, and ARC\_V2A~\cite{clauser2025arc}.
The physical meshes are first mapped to the unit disk, predictions are made in the canonical domain, and the resulting temperature fields are mapped back to physical coordinates for comparison with the M3D-C1 benchmarks.
The predicted temperature updates reproduce the dominant spatial structure across all four devices, with mean relative \(L^2\) errors in \(\Delta T_e\) of \(2.2\)--\(3.4\%\) across the geometry-specific validation sets.

To test whether the canonical-domain recast depends strongly on the neural-operator backbone, we evaluate both Fourier neural operators (FNOs) and localized neural operators (LocalNOs), each at two model scales.
All four models use exactly the same canonical-domain inputs, and differ only in the neural-operator architecture and model capacity.
The medium models use 32 Fourier modes and hidden width 64, whereas the large models use 48 modes and hidden width 128.
For the architecture comparison, the same \(20\) validation samples from each device geometry are evaluated with all four models.
Because the full \(T_e\) field is dominated by its background profile, the increment \(\Delta T_e\) provides a more discriminating comparison of the learned update.
Table~\ref{tab:si-tokamak-architecture} summarizes the corresponding \(\Delta T_e\) errors.
Increasing model capacity improves both FNO and LocalNO, and both architecture families achieve low error using the same equation-recast representation.
LocalNO-L gives the lowest error for all four device geometries, with mean relative \(L^2\) errors between \(2.2\%\) and \(3.4\%\), and is therefore used for the main-text results.
Importantly, the successful use of both FNO and LocalNO with the same recast inputs shows that the canonical-domain formulation is not tied to a specific neural-operator backbone.

\begin{table}[htbp]
    \centering
    \footnotesize
    \setlength{\tabcolsep}{8pt}
    \renewcommand{\arraystretch}{1.12}
    \caption{
    \textbf{Neural-operator architecture ablation for the Tokamak benchmark.}
    Mean masked relative \(L^2\) error on the electron-temperature increment
    \(\Delta T_e\), evaluated on the same \(20\) validation samples from each device geometry.
    Lower values indicate better accuracy.
    }
    \label{tab:si-tokamak-architecture}
    \begin{tabular}{lccccc}
        \toprule
        Model
        & C-Mod
        & C-Mod\_FD
        & SPARC
        & ARC\_V2A
        & Mean \\
        \midrule

        FNO-M
        & \(6.49\%\)
        & \(7.75\%\)
        & \(6.74\%\)
        & \(10.48\%\)
        & \(7.86\%\) \\

        FNO-L
        & \(3.68\%\)
        & \(3.04\%\)
        & \(2.64\%\)
        & \(3.82\%\)
        & \(3.30\%\) \\

        LocalNO-M
        & \(5.63\%\)
        & \(6.29\%\)
        & \(5.60\%\)
        & \(8.61\%\)
        & \(6.53\%\) \\

        LocalNO-L
        & \(\mathbf{2.45\%}\)
        & \(\mathbf{2.22\%}\)
        & \(\mathbf{2.60\%}\)
        & \(\mathbf{3.40\%}\)
        & \(\mathbf{2.67\%}\) \\

        \bottomrule
    \end{tabular}
\end{table}

\section{Discussion}
\label{sec:discussion}

Equation recast changes how the learning burden is allocated in parametric PDE modeling.
Rather than learning an entire parameter-indexed family of solution operators, the framework learns a single canonical inverse operator and handles analytically known operator variation through the governing equation.
For a target configuration, the operator difference is absorbed into an effective source, allowing the same canonical inverse to be reused across parameter configurations through a fixed-point iteration.
Equation recast is therefore an equation-level learning reformulation that can be combined with different operator-learning models.

The results collectively illustrate three main properties of this reformulation.
First, a neural operator trained at a single canonical configuration can support zero-shot extrapolation over parameter ranges substantially broader than those represented during training, as demonstrated by the two-parameter ADR and nonlinear Navier--Stokes problems.
Second, the canonical representation enables sparse and heterogeneous data generated at different parameter configurations to be used jointly.
The reaction--diffusion example shows that incorporating data from a single off-canonical configuration improves predictions over a broader neighboring regime, while the Navier--Stokes ablations further demonstrate the benefits of combining data from scattered parameter configurations at larger training budgets.
Each parameter configuration contributes a distribution of effective sources rather than information confined to a single point in parameter space, allowing heterogeneous data to enrich the canonical learning problem.
Third, the fixed-point iteration provides an internal indicator of potentially unreliable predictions.
The Helmholtz example shows that convergence slows and eventually fails near resonance, reflecting the loss of a stable iterative solution pathway from the chosen canonical operator.
Although prediction accuracy still depends on the learned canonical inverse, slow convergence or divergence can provide a warning when extrapolation becomes unreliable.

The high-fidelity tokamak application extends the same formulation to data that vary in both plasma coefficients and device geometry.
Harmonic mapping transforms the physical domains into a shared canonical domain, where geometry dependence enters analytically through Jacobian-dependent metric terms.
These geometry-induced operator differences and plasma-coefficient variations are incorporated into the same canonical representation, allowing simulation data from substantially different devices to train a single canonical-domain neural operator rather than separate device-specific models.
This is particularly relevant to high-fidelity MHD surrogate development, where simulation data are expensive and fragmented across operating conditions and device geometries.

The shift from dense parameter-space sampling toward canonical source--solution coverage raises an open question of how to characterize the induced effective-source distribution and construct training data representative of the parameter variations of interest.
A potential advantage of this shift is greater flexibility in generating and integrating training data.
Manufactured solutions allow additional source--solution pairs to be constructed by evaluating the known canonical operator on prescribed solution fields, which is often substantially cheaper than solving the corresponding high-fidelity PDE.
Heterogeneous data generated at scattered parameter configurations can also be recast into the same canonical representation and jointly contribute to source-space coverage, without requiring the learned model to infer parameter dependence from sparsely distributed samples.
An important direction for future work is therefore to characterize the relevant canonical source--solution distribution and develop efficient strategies to improve its coverage.

Taken together, these results suggest a complementary direction to increasingly expressive parametric operator models.
When variation across a PDE family is already characterized by the governing equation, that structure need not be learned again solely from data.
Instead, a reusable surrogate can combine canonical operator learning with analytically derived operator differences and heterogeneous effective-source data.
This division of labor preserves the flexibility of neural operators while exploiting known PDE structure to support generalization across parameter configurations and geometries.
For scientific applications in which simulations are expensive and parameter coverage is sparse, equation recast offers a route toward more reusable and data-efficient neural PDE solvers.
\clearpage
\section*{Acknowledgments}

The authors gratefully acknowledge funding from Commonwealth Fusion Systems; the U.S. DOE FIRE Collaborative, “Mitigating Risks from Abrupt Confinement Loss (MiRACL),” under Contract No. DE-AC02-09CH11466; and the DOE Genesis Mission under Award No. DE-SC0026734.

\section*{Data availability}

The datasets generated and analyzed during this study, including the
advection-diffusion-reaction, reaction-diffusion, Helmholtz, and
Navier-Stokes benchmark datasets, the trained neural-operator weights for all
five benchmark problems including the four Tokamak models, and the saved
evaluation outputs underlying every reported figure and table, are openly
available in the Zenodo repository at \url{https://doi.org/10.5281/zenodo.22016990}.

\section*{Code availability}

All code needed to reproduce the results of this study is publicly available
under the MIT license at \url{https://github.com/QiyunCh/EquationRecast}.
The repository provides the reference solvers, dataset generators, training
scripts, equation-recast inference, and plotting scripts for all five benchmark
problems, together with a mapping from every figure and table in the manuscript
to the script that produces it.

\clearpage
\colorlet{apprevisioncolor}{black}
\newcommand{\apprev}[1]{\textcolor{apprevisioncolor}{#1}}
\newenvironment{apprevision}{\par\begingroup\color{apprevisioncolor}}{\par\endgroup}
\appendix
\renewcommand{\thesection}{\Alph{section}}
\renewcommand{\thesubsection}{\thesection.\arabic{subsection}}
\counterwithin*{equation}{section}
\counterwithin*{table}{section}
\counterwithin*{figure}{section}
\renewcommand{\theequation}{\thesection\arabic{equation}}
\renewcommand{\thetable}{\thesection\arabic{table}}
\renewcommand{\thefigure}{\thesection\arabic{figure}}
\makeatletter
\@ifundefined{theHsection}{}{\renewcommand{\theHsection}{\Alph{section}}}
\@ifundefined{theHequation}{}{\renewcommand{\theHequation}{\thesection\arabic{equation}}}
\@ifundefined{theHtable}{}{\renewcommand{\theHtable}{\thesection\arabic{table}}}
\@ifundefined{theHfigure}{}{\renewcommand{\theHfigure}{\thesection\arabic{figure}}}
\makeatother

\section*{Appendix}
\phantomsection
\addcontentsline{toc}{section}{Appendix}

\section{Advection--diffusion--reaction equation}
\label{app:adr}

\begin{apprevision}
The ADR benchmark uses the periodic domain \(x\in[0,1)\), with \(u(0)=u(1)\) and \(u'(0)=u'(1)\), and the canonical configuration \((\mathrm{Pe}^*,\mathrm{Da}^*)=(4,2)\) of Table~\ref{tab:equation_recast_cases}.
Source fields are sampled as periodic Gaussian random fields (GRFs) in Fourier space with spectrum
\(E(\kappa)\propto\exp[-(\ell\kappa)^2/2]\), using \(\ell=0.08\).
The zero mode is removed and each source is rescaled to \([-1,1]\).
Reference solutions are generated on a periodic grid with \(N=201\) using a Fourier spectral solver.

The canonical inverse is represented by a one-dimensional FNO with three Fourier layers, 64 modes, and width 64.
Of the \(1000\) canonical samples, \(900\) are used for training and \(100\) for validation.
The model is trained for \(1000\) epochs using Adam and mean-squared-error loss.

Inference starts from \(u^{(0)}=\mathcal{G}_N^*[S]\) and updates the effective source as
\begin{equation}
    S_{\mathrm{eff}}^{(m)}
    = S-(\mathrm{Pe}-\mathrm{Pe}^*)\frac{du^{(m)}}{dx}
       -(\mathrm{Da}-\mathrm{Da}^*)u^{(m)},
    \label{eq:adr-effective-source}
\end{equation}
with the derivative evaluated spectrally.
Aitken's \(\Delta^2\) relaxation is used, and the iteration terminates when the relative iterate change falls below \(10^{-5}\) or after \(100\) iterations.

The parameter scan uses unit spacing over \((\mathrm{Pe},\mathrm{Da})\in[1,20]^2\), giving \(400\) configurations.
At each configuration, the mean relative \(L^2\) error and iteration count are evaluated over \(20\) independent test sources from the training GRF distribution.
The canonical-configuration sensitivity study in Fig.~\ref{fig:Figure_s1} uses the same source distribution and training-set size; the reference points and extended scan are specified in Section~\ref{sec:results:adr}.
\end{apprevision}

\section{Reaction--diffusion equation}
\label{app:dr}

\begin{apprevision}
The reaction--diffusion benchmark is defined on \(x\in[0,L]\), with \(L=10\) and homogeneous Dirichlet conditions \(u(0)=u(L)=0\).
Source fields are sampled from a GRF with covariance
\(C(x,x')=\sigma^2\exp[-(x-x')^2/(2\ell^2)]\), using \(\ell=0.5\) and \(\sigma=1\), and each realization is rescaled to \([-1,1]\).
Reference solutions are generated on a uniform grid with \(N=201\) using second-order finite differences and a sparse direct solver.

Both models use a total data budget of \(1000\) samples.
Model~1 uses only samples at \(k^*=0.785\); Model~2 uses \(500\) canonical samples and \(500\) samples at \(k_{\mathrm{new}}=0.05\).
For a training pair \((S,u)\) generated at coefficient \(k\), the canonical input is
\begin{equation}
    S_{\mathrm{eff}}=S-(k-k^*)u,
    \label{eq:dr-training-recast}
\end{equation}
while \(u\) remains the target.

The models use a one-dimensional FNO with three Fourier layers, 64 modes, and width 64.
A fixed coordinate channel \(\xi\in[-1,1]\) accompanies the source input.
Each dataset is split into \(90\%\) training and \(10\%\) validation samples, and training uses Adam and mean-squared-error loss.

At inference, Eq.~\eqref{eq:dr-training-recast} is evaluated with the current solution estimate; Aitken's \(\Delta^2\) relaxation is used to update that estimate.
Both models are tested on the same \(20\) independent GRF sources at \(27\) reaction coefficients spanning \(k\in[0.01,10]\), including the canonical and added values.
The reported error is the mean relative \(L^2\) error against the finite-difference reference solutions.
\end{apprevision}

\section{Helmholtz equation}
\label{app:helmholtz}

\begin{apprevision}
The Helmholtz benchmark uses the domain and Dirichlet conditions stated in Section~\ref{sec:results:helmholtz}.
Source generation, the \(N=201\) finite-difference discretization, the sparse direct solver, and the FNO architecture and training settings follow Appendix~\ref{app:dr}.
Model~1 uses \(1000\) samples at \(k^*=0.785\); Model~2 uses \(500\) canonical samples and \(500\) samples at \(k_{\mathrm{new}}=1.099\).

For a training pair \((S,u)\) generated at wavenumber \(k\), the canonical effective source is
\begin{equation}
    S_{\mathrm{eff}}=-S-\bigl(k^2-(k^*)^2\bigr)u,
    \label{eq:helmholtz-training-recast}
\end{equation}
so that \(\mathcal{O}^*[u]=S_{\mathrm{eff}}\), consistent with Table~\ref{tab:equation_recast_cases}.
The implementation supplies \(-S_{\mathrm{eff}}\) to its positive-source input channel, while retaining \(u\) as the target.
During inference, the effective source is evaluated with the current solution estimate; Aitken's \(\Delta^2\) relaxation is used to update that estimate.

Both models are evaluated on the same \(20\) independent test sources over \(k\in[0.630,1.200]\), including the canonical and added training values and both sides of \(k_3\).
Mean relative \(L^2\) errors and mean recast iteration counts are reported in Fig.~\ref{fig:Figure3b}.
\end{apprevision}

\section{Two-dimensional steady Navier--Stokes equation}
\label{app:ns}

\begin{apprevision}
Equation~\eqref{eq:ns_vorticity} is solved on the periodic domain \([0,1)^2\) with zero-mean fields.
The velocity is recovered from \(\Delta\psi=-\omega\) as \(\mathbf{u}=(\partial_y\psi,-\partial_x\psi)\).
Reference solutions use a \(128\times128\) Fourier pseudo-spectral grid and the \(2/3\) de-aliasing rule.
Steady states are obtained by IMEX pseudo-time integration with \(\Delta t=0.2\), terminating at a relative PDE residual below \(10^{-6}\) or after \(2500\) steps.
Forcing fields are generated by filtering Fourier-space white noise with power spectrum proportional to \(|k|^{-2}\), restricted to \(2\leq|k|\leq21\).
Each realization has zero mean and unit standard deviation.

The canonical and conditioned models use the datasets described in Section~\ref{sec:results:ns}, each split into \(90\%\) training and \(10\%\) validation samples.
All models share a two-dimensional FNO backbone with four Fourier layers, \(32\times32\) retained modes, and hidden width 64.
The canonical model maps \(S\) to \(\omega\); the conditioned models additionally receive \(\mathrm{Re}\) as a spatially constant input channel.
The parametric FNO uses supervised mean-squared-error loss, and PINO includes an additional PDE-residual term.
The data-budget ablation in Fig.~\ref{fig:si-budget-vs-Re} uses the same backbone and reference solver, with datasets and recast inputs defined in the main text.

For recast inference, the Laplacian in the effective-source correction is evaluated spectrally, and the normalized source is passed to the canonical FNO.
Aitken's \(\Delta^2\) relaxation is the default update.
All models are tested on the same \(20\) independent forcing fields over \(\mathrm{Re}\in[50,400]\), with a reference solution recomputed at each Reynolds number.
In addition to relative \(L^2\) error, we report the normalized target-PDE residual
\begin{equation}
    \varepsilon_{\mathrm{PDE}}
    = \frac{\left\|\mathbf{u}(\omega)\cdot\nabla\omega
                    -\mathrm{Re}^{-1}\Delta\omega-S\right\|_2}{\|S\|_2}.
    \label{eq:ns2d-pde-residual}
\end{equation}
\end{apprevision}

\subsection{Inference behavior and iterative-solver ablation}
\label{sec:si-ns-inference}

\begin{figure}[htbp]
    \centering
    \includegraphics[width=0.98\textwidth]{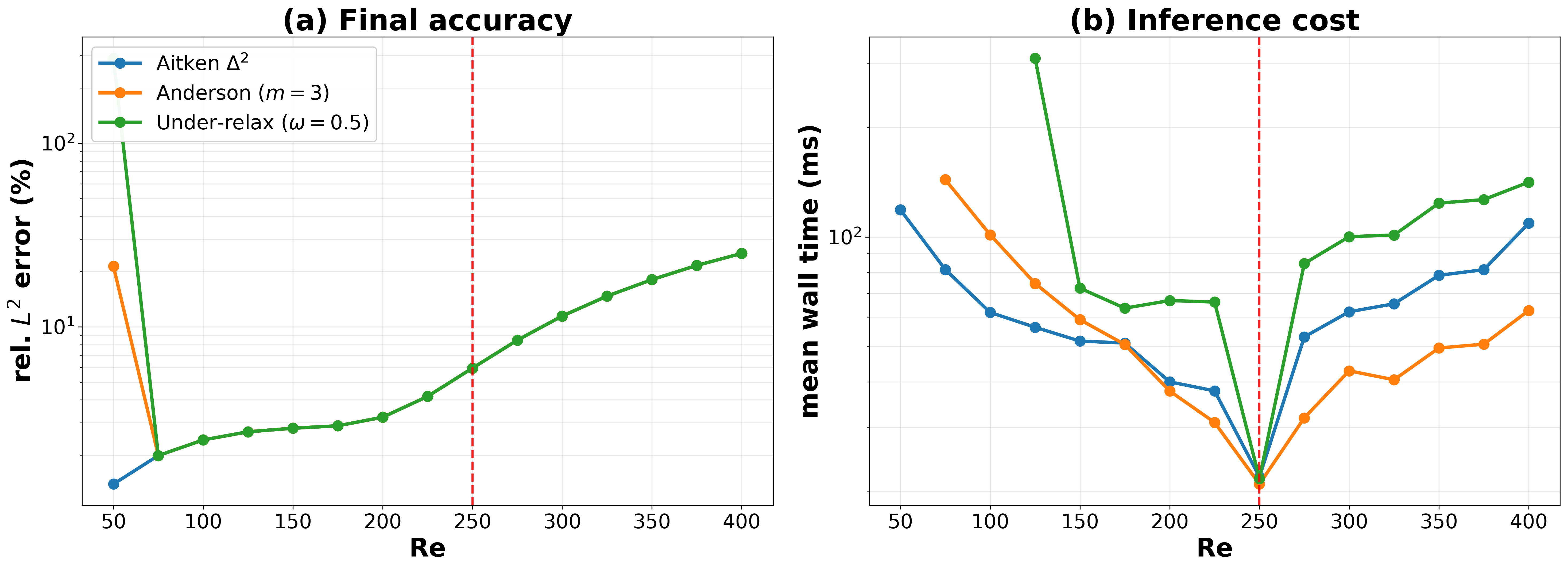}
    \caption{
    \textbf{Effect of the iterative update scheme on Navier--Stokes equation recast.}
    Final mean relative \(L^2\) error \textbf{(a)} and mean wall-clock time \textbf{(b)} as functions of Reynolds number for Aitken \(\Delta^2\) relaxation, Anderson mixing (\(m=3\)), and fixed under-relaxation (\(\alpha=0.5\)).
    Results are averaged over \(20\) test sources at each Reynolds number.
    Non-converged cases are excluded from the timing averages.
    The canonical configuration is \(\mathrm{Re}^*=250\).
    The corresponding iteration counts are shown in Fig.~\ref{fig:ns-convergence}(b).
    }
    \label{fig:si-damping}
\end{figure}

\begin{apprevision}
We compare Aitken's \(\Delta^2\) relaxation, Anderson mixing with history length 3, and fixed under-relaxation with \(\alpha=0.5\) over \(15\) Reynolds numbers in \([50,400]\) and \(20\) sources per Reynolds number, giving \(300\) cases.
All schemes use the same canonical FNO and initial iterate \(\omega^{(0)}=\mathcal{G}_N^*[S]\).
Each source is iterated individually until \(\|\omega^{(m+1)}-\omega^{(m)}\|_2/\|\omega^{(m)}\|_2<10^{-5}\) or a limit of \(200\) iterations, with wall-clock time measured per source on an NVIDIA RTX 4090 GPU.
Table~\ref{tab:si-iteration} reports aggregate results; Fig.~\ref{fig:si-damping} shows errors and timings by Reynolds number, complementing the iteration counts in Fig.~\ref{fig:ns-convergence}(b).
\end{apprevision}

\begin{table}[h]
    \centering
    \footnotesize
    \setlength{\tabcolsep}{7pt}
    \renewcommand{\arraystretch}{1.12}
    \caption{
    \textbf{Relaxation-scheme ablation for Navier--Stokes equation recast.}
    Results are aggregated over \(15\) Reynolds numbers in
    \(\mathrm{Re}\in[50,400]\) and \(20\) test sources per Reynolds number.
    Iteration counts and wall-clock times are averaged over converged cases.
    }
    \label{tab:si-iteration}
    \begin{tabular}{lcccc}
        \toprule
        Scheme
        & Mean iterations
        & Mean time (ms)
        & Mean rel.\ \(L^2\)
        & Non-converged \\
        \midrule
        Aitken \(\Delta^2\)
        & \apprev{\(10.4\)}
        & \apprev{\(64.8\)}
        & \apprev{\(\mathbf{8.5\%}\)}
        & \(\mathbf{0/300}\) \\

        Anderson (\(m=3\))
        & \apprev{\(\mathbf{8.0}\)}
        & \apprev{\(\mathbf{50.7}\)}
        & \apprev{\(9.8\%\)}
        & \apprev{\(39/300\)} \\

        Fixed relaxation (\(\alpha=0.5\))
        & \apprev{\(17.5\)}
        & \apprev{\(93.0\)}
        & \apprev{\(27.7\%\)}
        & \apprev{\(75/300\)} \\
        \bottomrule
    \end{tabular}
\end{table}

\paragraph{Inference cost.}
\begin{apprevision}
Table~\ref{tab:si-timing} compares per-source wall-clock times at matched prediction error.
At each Reynolds number, the target is the median over \(20\) sources of the minimum relative \(L^2\) error reached within \(80\) Aitken-relaxed iterations in Fig.~\ref{fig:ns-convergence}(a).
For recast inference, all \(20\) sources are batched on the GPU.
For each source, the cumulative time to first reach within \(5\%\) of its plateau error is divided by the batch size; the minimum over two batch runs is used, and the median over sources is reported.

The pseudo-spectral solver is run on the GPU (PyTorch) and CPU (NumPy), sweeping relative residual tolerances from \(10^{-2}\) to \(10^{-6}\).
Errors are measured against a reference solution converged to tolerance \(10^{-7}\), and the loosest tolerance meeting the target error is selected.
Each solve is timed as the median of \(10\) repetitions after \(5\) warm-up runs, with the final median taken over \(5\) sources.
All timings use a \(128\times128\) grid on the same machine with an NVIDIA RTX 4090 GPU.
At \(\mathrm{Re}^*=250\), the recast plateau is reached at the initial iterate, so inference requires a single canonical FNO evaluation.
\end{apprevision}

\begin{table}[htbp]
    \centering
    \footnotesize
    \setlength{\tabcolsep}{8pt}
    \renewcommand{\arraystretch}{1.12}
    \caption{
    \textbf{Inference cost at matched prediction accuracy for the Navier--Stokes benchmark.}
    Wall-clock time per sample required by equation recast with Aitken \(\Delta^2\) relaxation and by the pseudo-spectral numerical solver on GPU and CPU to reach the same relative \(L^2\) error.
    \apprev{Recast times are reported per source from a GPU batch of \(20\); solver times use individual solves at the loosest residual tolerance meeting the target error.
    Results are medians over \(20\) sources for recast and \(5\) for the solver, using a \(128\times128\) grid on the same machine with an NVIDIA RTX 4090 GPU.
    The timing protocol and target-error definition are given in the text.}
    }
    \label{tab:si-timing}
    \begin{tabular}{lccccc}
        \toprule
        & \(\mathrm{Re}=50\) & \(\mathrm{Re}=150\) & \(\mathrm{Re}=250\) & \(\mathrm{Re}=350\) & \(\mathrm{Re}=400\) \\
        \midrule
        Target rel.\ \(L^2\) error (\%) & \apprev{\(1.30\)} & \apprev{\(2.63\)} & \apprev{\(5.25\)} & \apprev{\(16.58\)} & \apprev{\(23.49\)} \\
        Equation recast (ms) & \apprev{\(11.2\)} & \apprev{\(5.1\)} & \apprev{\(1.0\)} & \apprev{\(5.1\)} & \apprev{\(5.1\)} \\
        Numerical solver, GPU (ms) & \apprev{\(18.4\)} & \apprev{\(14.0\)} & \apprev{\(16.2\)} & \apprev{\(32.4\)} & \apprev{\(62.1\)} \\
        Numerical solver, CPU (ms) & \apprev{\(25.0\)} & \apprev{\(25.3\)} & \apprev{\(50.3\)} & \apprev{\(96.2\)} & \apprev{\(126.1\)} \\
        \bottomrule
    \end{tabular}
\end{table}

\section{Multi-device Tokamak MHD simulation}
\label{app:tokamak}

\begin{apprevision}
The Tokamak benchmark uses electron-temperature evolution data from high-fidelity M3D-C1 simulations for Alcator C-Mod, C-Mod\_FD, SPARC, and ARC\_V2A.
The surrogate predicts the electron-temperature update, with the remaining simulated physics entering through the coefficient fields \(\mathbf{p}=(n_e,\sigma_e,\kappa)\) and the source \(Q_{\mathrm{tot}}\) in Eq.~\eqref{eq:tokamak_physical}.
Advective transport is not included in this benchmark.
\end{apprevision}

\paragraph{Mapping to the canonical domain.}
\begin{apprevision}
Each physical device domain is mapped harmonically to a shared unit disk with canonical coordinates \(\boldsymbol{\xi}\).
For the inverse map \(\mathbf{x}=\phi(\boldsymbol{\xi})\), the Jacobian is \(\mathrm{K}=\partial\mathbf{x}/\partial\boldsymbol{\xi}\), with determinant \(|\mathrm{K}|\).
Gradients transform as \(\nabla=\mathrm{K}^{-\mathsf{T}}\nabla_{\boldsymbol{\xi}}\), and the divergence of a vector field \(\mathbf{F}\) becomes
\begin{equation}
    \nabla\cdot\mathbf{F}
    =
    \frac{1}{|\mathrm{K}|}\,
    \nabla_{\boldsymbol{\xi}}\cdot
    \left(|\mathrm{K}|\,\mathrm{K}^{-1}\mathbf{F}\right).
    \label{eq:si-tokamak-divergence}
\end{equation}
Applying this transformation to the heat flux gives the diffusion operator \(\mathcal{L}_{\kappa,\mathrm{K}}\) in Eq.~\eqref{eq:tokamak_conduction_operator}.
The map and its Jacobian are computed once for each device and reused during preprocessing and inference.
\end{apprevision}

\paragraph{Effective-source construction.}
\begin{apprevision}
Alcator C-Mod provides the reference geometry \(\mathrm{K}^*\).
The reference coefficient fields \(\tilde{\mathbf{p}}^*\), temperature \(\widetilde{T}_e^*\), and time step \(\Delta t^*\) are taken from one representative C-Mod frame, defining the canonical increment operator in Eq.~\eqref{eq:tokamak_onestep_operator}.
All one-step pairs use \(\Delta t^{n+1}=\Delta t^*\).
To construct the model input, we set the increment to zero in the operator difference of Eq.~\eqref{eq:tokamak_recast} and evaluate the case-dependent source and coefficients from the state at \(t^n\).
This gives the effective-source approximation
\begin{equation}
\begin{aligned}
    \widetilde{S}_{\mathrm{eff}}^{\,n+1,(0)}
    &\simeq
    \widetilde{Q}_{\mathrm{tot}}^{\,n}
    - \frac{\tilde{\sigma}_e}{\tilde{n}_e}\,\widetilde{T}_e^{\,n}
    + \frac{\gamma-1}{\tilde{n}_e}\,
      \mathcal{L}_{\kappa,\mathrm{K}}\!\left[\widetilde{T}_e^{\,n}\right]
    \\
    &\quad
    + \frac{\tilde{\sigma}_e^*}{\tilde{n}_e^*}\,\widetilde{T}_e^*
    - \frac{\gamma-1}{\tilde{n}_e^*}\,
      \mathcal{L}_{\kappa,\mathrm{K}}^*\!\left[\widetilde{T}_e^*\right],
\end{aligned}
\label{eq:si-tokamak-effective-source}
\end{equation}
where \(\mathcal{L}_{\kappa,\mathrm{K}}^*\) is evaluated with \(\tilde{\kappa}^*\) and \(\mathrm{K}^*\).
The first three terms give the temperature evolution rate at the current state; the last two form a fixed reference correction.
The effective source in Eq.~\eqref{eq:si-tokamak-effective-source} is evaluated at zero increment and is therefore an explicit approximation to the increment-dependent effective source in Eq.~\eqref{eq:tokamak_recast}.
The source and coefficient fields are additionally evaluated at the current time level \(t^n\), rather than at the next time level \(t^{n+1}\) used in the backward-Euler formulation; this is a separate approximation from setting the increment to zero.
\end{apprevision}

\paragraph{Training pairs and prediction.}
\begin{apprevision}
For consecutive simulation states separated by \(\Delta t^*\), the implemented surrogate is trained on the current temperature field and the approximate effective source,
\[
    \left(\widetilde{T}_e^{\,n},\widetilde{S}_{\mathrm{eff}}^{\,n+1,(0)}\right),
\]
with the full-step M3D-C1 temperature increment as the supervision target,
\begin{equation}
    \Delta\widetilde{T}_e^{\,n+1}
    = \widetilde{T}_e^{\,n+1}-\widetilde{T}_e^{\,n}.
    \label{eq:si-tokamak-target}
\end{equation}
This conditional one-step mapping is evaluated once per step as in Eq.~\eqref{eq:tokamak_one_step}, without iterating the recast equation to convergence.
The predicted increment is added to the current temperature.
Using the increment as the target separates the temperature change from the dominant background profile.

After removing states affected by numerical artifacts and initialization transients, the four device datasets contain \(4{,}208\) one-step pairs in total, with \(15\%\) held out for validation.
All four geometries contribute to training.
Fields are represented on a \(256\times256\) canonical-disk grid, and the training loss is evaluated over valid mapped pixels using a binary domain mask.
Further simulation, preprocessing, and implementation details are provided with the open-source code and simulation references.
\end{apprevision}

\clearpage

\bibliographystyle{apsrev4-2}

\makeatletter
\if@filesw\immediate\write\@auxout{\string\citation{apsrev42Control}}\fi
\makeatother
\bibliography{apsrev42Control,refs}

\end{document}